\documentclass{article}
\usepackage{arxiv}

\usepackage[utf8]{inputenc} 
\usepackage[T1]{fontenc}    
\usepackage{url}            
\usepackage{booktabs}       
\usepackage{amsfonts}       
\usepackage{nicefrac}       
\usepackage{microtype}      
\usepackage{lipsum}		
\usepackage{graphicx}
\usepackage{natbib}

\usepackage{doi}
\usepackage{multirow}
\usepackage{multicol}
\usepackage[frozencache,cachedir=.]{minted}
\usepackage{alltt}
\usepackage{longtable}
\usepackage{amsmath}
\usepackage[normalem]{ulem}
\usepackage[dvipsnames]{xcolor}
\usepackage{tcolorbox}   
\usepackage{linguex}

\LTcapwidth=\textwidth

\title{
    The Lucie-7B LLM and the Lucie Training Dataset:
    Open resources for multilingual language generation
}
\author{
    Olivier Gouvert (1)\thanks{(1) principal contributors (Olivier: model training at all stages, data preparation; Julie: oversight of data preparation and mixes, day-to-day management; Jérôme: tokenizer training, model training first phase and project technical lead); (2) contributors to model training; (3) principal contributor to instruction tuning; (4) team manager.} \\  
	LINAGORA\\
    Toulouse, France\\
	\texttt{ogouvert@linagora.com} 
    	\And
    Julie Hunter (1)
    \\
	LINAGORA\\
	Toulouse, France \\
	\texttt{jhunter@linagora.com} \\
	\And
    Jérôme Louradour (1) %
    \\
	LINAGORA\\
	Toulouse, France \\
	\texttt{jlouradour@linagora.com} \\
    	\And
    Christophe Cerisara (2)\\
	LORIA\\
	Paris, France \\
	\texttt{christophe.cerisara@loria.fr} \\
        \And
    Evan Dufraisse (2)\\
	CEA List\\
	Palaiseau, France \\
	\texttt{evan.dufraisse@cea.fr} \\
    	\And
    Yaya Sy (2)\\
	LORIA\\
	Paris, France \\
	\texttt{yaya.sy@loria.fr} \\
        \And
    Laura Rivière (3)\\
	LINAGORA\\
	Toulouse, France \\
	\texttt{lriviere@linagora.com} \\
        \And
    Jean-Pierre Lorré (4) \\
	LINAGORA\\
	Toulouse, France \\
	\texttt{jplorre@linagora.com} \\
        \And
    OpenLLM-France community \\
    \texttt{contact@openllm-france.fr}
}

\date{January 2025}

\renewcommand{\headeright}{}
\renewcommand{\undertitle}{}

\newcommand{\usedapache}{$^{(\dagger)}$}

\DeclareUnicodeCharacter{2503}{|}
\begin{document}

\maketitle

\begin{abstract}
    We present both the Lucie Training Dataset and the Lucie-7B foundation model, open resources created by the OpenLLM-France community.\footnote{The OpenLLM France Community is a community of people interested in all aspects of the development of open language models and is organized mainly around the French section of a Discord server now called OpenLLM Europe.} The Lucie Training Dataset is a multilingual collection of textual corpora centered around French, and designed to offset the anglo-centric biases found in many datasets for large language model pretraining. Its French data is pulled not only from the traditional web data sources, but also from French cultural heritage documents, filling an important gap in modern datasets. Beyond French, which makes up the largest share of the data, we added documents to support several other European languages, including English, Spanish, German, and Italian.
    Apart from its value as a resource for French language and culture, an important feature of this dataset is that it prioritizes data rights by minimizing copyrighted material. In addition, building on the philosophy of past open projects, it is redistributed\footnote{A single dataset, accounting for 0.2\% of the final dataset, could not be redistributed as explained in Section \ref{sec:data} but its content is described on Hugging Face.} in the form used for training and its processing is described on Hugging Face and GitHub. 
    The Lucie-7B foundation model is trained on equal amounts of data in French and English---roughly 33\% for each---in an effort to better represent cultural aspects of French-speaking communities.
    We also describe two instruction fine-tuned versions of Lucie-7B, Lucie-7B-Instruct-v1.1 and Lucie-7B-Instruct-human-data, which we release as demonstrations of the foundation model in use. These models achieve promising results compared to state-of-the-art models, demonstrating that an open approach prioritizing data rights can still deliver strong performance.
    We see these models as an initial step toward developing more performant, and aligned models in the near future. The Lucie-7B resources, like the dataset, are all open. Model weights for Lucie-7B and the Lucie instruct models, along with intermediate checkpoints for the former, are likewise published on Hugging Face, while model training and data preparation code is freely available on GitHub. This makes Lucie-7B one of the first OSI compliant language models according to the new OSI definition.\footnote{Open Source AI Definition \url{https://opensource.org/ai/open-source-ai-definition}} 
\end{abstract}



\section{Introduction}
When we set out to create Lucie-7B, one of our principal goals was to offset the English-centered bias found in the training sets of most mainstream large language models (LLMs) such as Meta's Llama models. Poor representation of a language can lead not only to reduced language modeling performance for that language, but what's more, a model that has seen scant data from a certain language is likely to have received scant information about cultures associated with that language. When faced with a question about history, religion, social practices or even everyday activities like cooking, the model is more likely to provide an answer suited to the language on which it is predominantly trained, which can lead to unsatisfying results for members of other language communities. For these reasons, together with the fact that the Lucie team was based in France, we decided early on that Lucie-7B would see at least as much French data as English. 

A second priority was to ensure that our data collection practices were respectful of the rights of the data producers. This includes abiding by copyright and intellectual property restrictions as well as eliminating personally identifying information where possible. An immediate consequence of this choice was that we needed to be careful in using data collected from the web. Initially, our plans were to rely only on sources, such as Wikipedia, that are well known to provide clean, high-quality data. The reality of the situation, however, is that the vast majority of LLM training data comes from the web and this data provides a level of quantity and diversity that is hard to reach through other means. We ultimately decided to gently relax our prohibition of general web data,  though limiting it remained a guiding principle throughout our data collection process. 

Finally, building on the open-source philosophy of LINAGORA, the founding partner of the OpenLLM France community, a third central motivation for our project was to further growing efforts to shed light on LLM development. In fact, the name ``Lucie'' is derived from the Latin word for ``light,'' reflecting our desire make every aspect of our process visible to the public.  From the beginning, we committed to sharing all information about our training data, including publication of the training datasets where possible, model weights, evaluation results, and all code used for training and data preparation. 

What we did not set out to do was to create an all-purpose chat assistant in the style of ChatGPT. Not only is ChatGPT allegedly hundreds of times larger than Lucie, but it, like all of the best performing models nowadays, has undergone multiple phases of fine-tuning, including alignment through methods like Reinforcement Learning with Human Feedback (RLHF) or Direct Preference Optimization (DPO), with large amounts of high-quality data. Collecting the data needed for these methods was beyond our resources in terms of time and human labor for this particular project. 
In the context of a recently begun research grant that carries the same name as the OpenLLM-France community,\footnote{We gratefully acknowledge support from the BPI funded project OpenLLM-France.} however, we are working with a larger, dedicated group of partners to build aligned versions of Lucie-7B. The resources described in this paper are thus the first (major) steps in a larger research program. We go ahead and release them in the spirit of open-source collaboration so that those who wish can begin to use the resources for their own purposes. 

This paper describes in detail the Lucie Training Dataset and the Lucie-7B models, including the foundation model and the two associated instruction-tuned models released as demonstrations. The Lucie Training Dataset, most of which has been republished on Hugging Face,\footnote{\url{https://huggingface.co/datasets/OpenLLM-France/Lucie-Training-Dataset}}  is a collection of texts from diverse sources in French, English, Spanish, German and Italian as well as various programming languages. The majority of its data comes from French; it is in fact one of the biggest collections of French text data that has been preprocessed for LLM training. We provide further details in Section \ref{sec:data}. Lucie-7B, while proficient in all of the natural languages from the Lucie Training Dataset, is designed to be particularly competent at generating French and English: roughly one third of its training data is in English and another third is in French. It is presented in Section \ref{sec:model}, where we describe tokenizer training as well as the three stages of pretraining carried out to create the final model. Model weights, including weights for regular intermediate checkpoints, are available on Hugging Face.\footnote{\url{https://huggingface.co/OpenLLM-France/Lucie-7B}}  Code for training and data preparation is openly accessible on GitHub.\footnote{\url{https://github.com/OpenLLM-France/Lucie-Training}} Finally, in Section \ref{sec:instruction}, we present the two instruction models, Lucie-7B-Instruct-v1.1 and Lucie-7B-Instruct-human-data,\footnote{\url{https://huggingface.co/OpenLLM-France/Lucie-7B-Instruct-v1.1} and \url{https://huggingface.co/OpenLLM-France/Lucie-7B-Instruct-human-data}} providing information on the datasets used to train them and the final evaluations. We turn first, however, to a review of related efforts to push for open data and open LLM development.


\section{Related open initiatives}
The BigScience Workshop\footnote{\url{https://bigscience.huggingface.co/}} was the first large-scale open-source collaboration aiming at training open foundation models, producing the Bloom LLM family of models as well as openly distributing all associated resources, including the ROOTS training dataset~\citep{roots}, scripts and working technical reports~\citep{bloom_short}.
This initiative involved more than 1,000 researchers from 60 countries and 250 institutions.
The Bloom models were trained on the A100 partition of the Jean Zay supercomputer.\footnote{The Jean ZAy cluster, based in France, is operated by GENCI and managed by the CNRS/IDRIS  \url{http://www.idris.fr/jean-zay/}.}
BigScience has been an inspiration for all following open-source LLM initiatives, including ours and those described below.

EleutherAI\footnote{\url{https://eleuther.ai/}} was another early actor in the push towards open foundation models and LLM pretraining data. It is a non-profit institute focusing on large-scale artificial intelligence research that made model and data sharing a central priority through its Pythia suite of open models and its dataset, the Pile, which combined a variety of web-based sources and domain-specific datasets in English~\citep{pythia}.  
The Allen Institute for AI\footnote{\url{https://allenai.org/}} is an active player in developing novel approaches for pretraining and has openly shared model weights and checkpoints for its OLMO models as well as training code~\citep{olmo}. They likewise share their pretraining dataset DOLMA~\citep{dolma} along with data and code for instruction tuning and evaluation\footnote{\url{https://github.com/allenai/OLMo-Eval}}. Like the Pile, DOLMA is heavily English-focused.

The CroissantLLM~\citep{croissant} project is highly complementary to that of Lucie-7B in that  CroissantLLM is focused on French and English, having seen equal amounts of data in these two languages. In addition to the weights of this 1.3 billion parameter model, training data is shared on Hugging Face and code is released on GitHub. 

The above initiatives have benefited from massive efforts to share general web-based data. The Common Crawl foundation\footnote{\href{https://commoncrawl.org/}{https://commoncrawl.org/}} regularly crawls the web to pick up new material, paying attention to respect opt-out choices from url holders. The OSCAR project~\citep{suarez2019asynchronous} shares its pipelines for classifying and cleaning Common Crawl dumps and freely shares the resulting datasets. C4 ~\citep{raffel2020} is another effort to clean Common Crawl data. RedPajama~\citep{redpajama} is a collection of web-based data including C4, Common Crawl data that has been processed with the CCNet pipeline and tagged with CCNet labels \citep{wenzek-etal-2020-ccnet}, and a handful of specialized web-sources such as Arxiv, Wikipedia and StackExchange. RefinedWeb ~\citep{penedo2023refinedweb} is the result of efforts to clean and filter web data for the Falcon LLM \citep{almazrouei2023falconseriesopenlanguage}. FineWeb is a curated web dataset filtered using a Bert-based embedder with a regression head, trained on Llama3-70B rating annotations, designed to identify high-quality web content~\citep{finewebedu}.

Open projects need platforms for distributing their data and models. Hugging Face,~\footnote{\url{https://huggingface.co}} which began as an initiative of The BigScience Workshop, has become the worldwide reference place to store and distribute open LLM resources, including model weights, code and datasets, with its Hugging Face Hub.~\footnote{\url{https://huggingface.co/docs/hub/index}} All of our Lucie-related resources are also available on the Hub. Our training code is available through GitHub.  

The OpenLLM France initiative benefits from, and builds upon, the open source and open data efforts described above. Its goal is to develop LLM training datasets and models that cater to French-speakers and, more generally, to Europeans in order to reduce European dependence on American and Chinese actors and push a truly open approach to AI development. Our first project yielded the 
Claire French Dialogue Dataset~\citep{claire_fr}, an open dataset distributed through Hugging Face. It is a collection of dialogues in French taken either from transcripts of conversations or from theater plays. This was later followed by its English counterpart, the Claire English Dialogue Dataset.  The Claire models are causal language models that result from continuing the training of Falcon~\citep{almazrouei2023falconseriesopenlanguage} and Mistral~\citep{jiang2023mistral} on the Claire French Dialogue Dataset with the aim of augmenting the representation of the French language and of French-speaking cultures in the resulting LLMs.


\section{The Lucie Training Dataset}\label{sec:data}

The Lucie Training Dataset is a curated collection of texts
in English, French, German, Spanish and Italian culled from a variety of sources including: web data, video subtitles, academic papers,
digital books, newspapers, and magazines. It also contains samples of diverse programming languages to  boost the reasoning capacity of models trained on this data. The majority of the texts in the corpus are in French, in an effort to minimize  anglo-centric cultural biases. 

In Section \ref{sec:datasets}, we introduce each of the subcorpora included in the Lucie Training Dataset and indicate their source and license information. All newspapers, monographies, magazines and legislative documents, as well as most books, are in the public domain
(which depends on the author's date of death and the country of publication). Other data are published with permissive licenses (e.g., CC BY or CC BY-SA) and in very rare cases, CC BY-NC-SA.   In Section \ref{sec:composition}, we breakdown the Lucie Training Dataset by category and show the contribution of each subcategory in terms of tokens to both the overall, raw dataset and to the final training dataset, for which certain datasets were upsampled.  Section \ref{sec:data-preparation} describes the various data-preprocessing approaches that were followed. For example, because of our emphasis on open data,  multiple subsets of the Lucie Training Dataset come from public archives, which leads to an increased number of texts processed by Optical Character Recognition (OCR). We describe our efforts to select higher quality data from these subsets. We also explain how we filtered web data in order to improve training and to stick as closely as possible to our principles regarding data rights and privacy. 

We note that one of the corpora used to train the Lucie-7B model is not distributed in the Lucie Training Dataset and so not included in the list below. This is because after training had begun, we discovered that the corpus, whose material was taken from the Persée platform,\footnote{https://www.persee.fr/} contained copyrighted material. The full list of urls used for the corpus can be found on the Hugging Face repository for the Lucie Training Dataset in the file ``persee\_metadata\_documents.csv''.\footnote{\url{https://huggingface.co/datasets/OpenLLM-France/Lucie-Training-Dataset/blob/main/metadata/persee_metadata_documents.csv}.  The corresponding url is \texttt{https://www.persee.fr/doc/\$ID} for each \texttt{ID} in the column \texttt{file\_id}.} The file ``persee\_metadata\_collections.csv''\footnote{\url{https://huggingface.co/datasets/OpenLLM-France/Lucie-Training-Dataset/blob/main/metadata/persee_metadata_collections.csv}.} gives statistics on document, word and character counts for the data grouped by collection. In all, the corpus contains a total of 3.25 billion words and 5.75 billion tokens, making up around 0.25\% of the raw corpus and 0.37\% of the tokens seen during training (as the data were seen twice during training as explained in Section \ref{sec:composition}).

\subsection{Data subset details}\label{sec:datasets}
For each subset of the Lucie Training Dataset, we identify the subset by the key we attributed to it in the final dataset and give a short description of the data contained in the subset. If we obtained the data from an external source, we cite our source. Where relevant and possible, we provide further information about how our source originally acquired the content and references to papers that the sources asked to have cited.

\subsubsection{AmendementsParlement}
AmendementsParlement is a collection of proposed amendments by the French parliament extracted from Regards Citoyens (\url{https://www.regardscitoyens.org/\#&panel1-4}).\footnote{CC BY-SA:  \url{https://www.regardscitoyens.org/mentions-legales/}} Documents contain the text of the proposed amendment, the name of the associated law, who voted on the amendment and what was decided.

\subsubsection{AmericanStories}
\uline{Source}: \url{https://huggingface.co/datasets/dell-research-harvard/AmericanStories}.\\ License: CC BY 4.0.

\begin{figure}[H]
    \centering
    \includegraphics[width=0.4\linewidth]{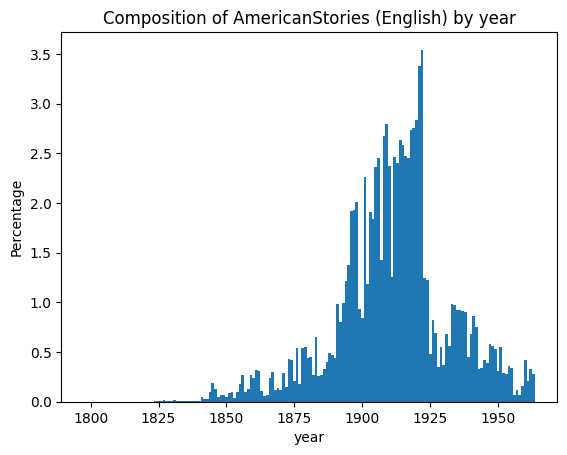}
    \caption{
        Distribution of documents by year in the AmericanStories dataset.
    }
    \label{fig:distribution_americanstories}
\end{figure}

AmericanStories~\citep{american_stories} is ``a collection of full article texts extracted from historical U.S. newspaper images. It includes nearly 20 million scans from the public domain Chronicling America\footnote{\url{https://www.loc.gov/collections/chronicling-america/about-this-collection/}; Open license:\\ \url{https://www.loc.gov/collections/chronicling-america/about-this-collection/rights-and-access/}} collection maintained by the Library of Congress. The dataset is designed to address the challenges posed by complex layouts and low OCR quality in existing newspaper datasets'' (from the Hugging Face data card). The composition details for statistics on documents by year are shown in Figure~\ref{fig:distribution_americanstories}. This dataset contains text retrieved through OCR.

\subsubsection{Claire (French and English)}
\begin{figure}[H]
    \centering
    \includegraphics[scale=0.7]{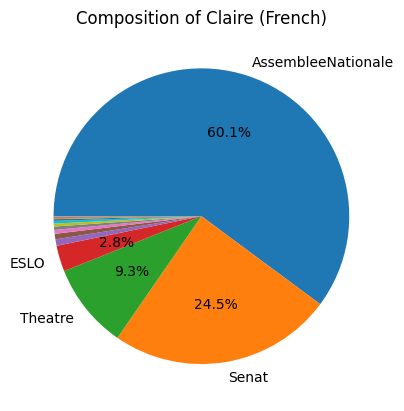} 
    \hfill
    \includegraphics[scale=0.7]{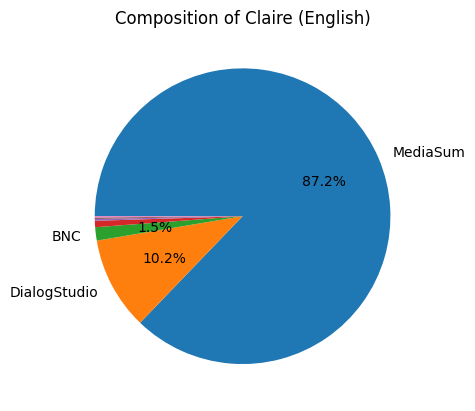} 
    \caption{
        Distribution of documents by type in the Claire French (left) and English (right) datasets.
    }
    \label{fig:distribution_claire}
\end{figure}

\uline{Sources}:
\begin{itemize}
    \item \url{https://huggingface.co/datasets/OpenLLM-France/Claire-Dialogue-French-0.1}.\\ License: CC BY-NC-SA 4.0.
    \item \url{https://huggingface.co/datasets/OpenLLM-France/Claire-Dialogue-English-0.1}.\\ License: CC BY-NC-SA 4.0.
\end{itemize}
The Claire datasets~\citep{claire_fr} are composed of transcripts of spoken conversations---including parliamentary proceedings, interviews, debates, meetings, and free conversations---as well as some written conversations from theater plays and written chats.
The dataset is designed to help downstream performance of models fine-tuned
for tasks requiring the comprehension of spontaneous spoken conversation, such as meeting summarization.
Each dialogue is split into speech turns,
and each speech turn is labeled with the name of the speaker or a unique identifier.
The composition details for the Claire French and English Datasets are shown in Figure~\ref{fig:distribution_claire}. For details on the subsets of the Claire datasets, see the respective data cards on Hugging Face.

\subsubsection{CroissantAligned}
\uline{Source}: \url{https://huggingface.co/datasets/croissantllm/croissant_dataset_no_web_data}\\
subset: \texttt{aligned\_36b};\footnote{\url{https://huggingface.co/datasets/croissantllm/croissant_dataset_no_web_data/tree/main/aligned_36b}} License: not specified
    
CroissantAligned~\citep{croissant} contains samples of parallel French/English (or English/French) data. Data extracted from OPUS\footnote{\url{https://opus.nlpl.eu/}} (99.6\% of the data in CroissantAligned) takes the form of sentences pairs, where one sentence is in French and the other is in English. OPUS pairs were passed through a custom pipeline designed to select the highest quality translation examples for CroissantLLM. The thesis abstract subset\footnote{License: ETALAB-Licence-Ouverte-v2.0} contains thesis abstracts paired with translations written by the thesis authors. The song lyrics are translated by contributors to \url{www.lacoccinelle.net}. Parallel data are used to boost the multilingual capabilities of models trained on them.

\subsubsection{DiscoursPublics}
DiscoursPublics is a collection of public speeches from the principal public actors in France including speeches from the French President starting from 1974 and from the Prime Minister and members of the government starting from 1980. The speeches are extracted from Vie Publique.\footnote{\url{https://www.vie-publique.fr/collection-discours-publics}; ETALAB-Licence-Ouverte-v2.0:\\ \url{https://www.vie-publique.fr/mentions-legales}}

\subsubsection{Europarl and EuroparlAligned}
\uline{Sources}:
\begin{itemize}
    \item \texttt{fr-en}, \texttt{es-en}, \texttt{it-en} parallel data: \url{https://www.statmt.org/europarl/v7/}. License: Open.
    \item \texttt{fr}, \texttt{en}, \texttt{de}, \texttt{es} monolingual data and \texttt{de-fr} parallel data: \url{https://www.statmt.org/europarl/v10/training-monolingual/}. License: Open.
\end{itemize}

The Europarl parallel corpus~\citep{europarl} is ``extracted from the proceedings of the European Parliament. It includes versions in 21 European languages: Romanic (French, Italian, Spanish, Portuguese, Romanian), Germanic (English, Dutch, German, Danish, Swedish), Slavik (Bulgarian, Czech, Polish, Slovak, Slovene), Finni-Ugric (Finnish, Hungarian, Estonian), Baltic (Latvian, Lithuanian), and Greek. The goal of the extraction and processing was to generate sentence aligned text for statistical machine translation systems'' (\url{https://www.statmt.org}).

\subsubsection{Eurovoc}
\uline{Source}: \url{https://huggingface.co/datasets/EuropeanParliament/Eurovoc}. License: EUPL 1.1\footnote{\url{https://huggingface.co/datasets/EuropeanParliament/Eurovoc}}

Eurovoc~\citep{eurovoc_1,eurovoc_2,eurovoc_3,eurovoc_4} is a collection of multilingual documents from the data repository of the Publications Office of the European Union annotated with Eurovoc labels. The corpus contains legal, policy-related, historical and organizational information about the EU. Dataset containing text retrieved through OCR and extracted from Cellar.\footnote{\url{https://op.europa.eu/en/web/cellar}; CC BY 4.0.}

\subsubsection{FineWebEdu}
\uline{Source}: \url{https://huggingface.co/datasets/HuggingFaceFW/fineweb-edu}. License: ODC-BY.

\begin{figure}[H]
    \centering
    \includegraphics[width=0.4\linewidth]{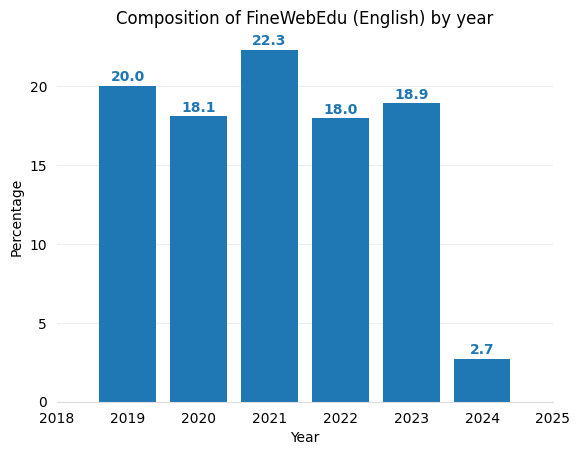}
    \caption{
        Distribution of documents by year in the FineWebEdu dataset.
    }
    \label{fig:distribution_finewebedu}
\end{figure}

FineWebEdu~\citep{finewebedu} is a 1.3 trillion token selection from FineWeb (\url{https://huggingface.co/datasets/HuggingFaceFW/fineweb}), which contains 15 trillion tokens of curated data from 96 Common Crawl dumps. Content in FineWebEdu has been selected by a custom designed classifier for its high-quality, educational content. The most recent crawl was done in October 2024.
Figure~\ref{fig:distribution_finewebedu} shows the distribution of documents by year in the FineWebEdu dataset.

\subsubsection{GallicaMonographies}
\uline{Source}: Corpus contributed by OpenLLM partners. A version is also published at \url{https://huggingface.co/datasets/PleIAs/French-PD-Books}. License: Public domain.

GallicaMonographies is a large collection of French monographies in the public domain made available through the French National Library (\url{https://gallica.bnf.fr/accueil/fr/content/accueil-fr?mode=desktop}) and extracted from Gallicagram.\footnote{\url{https://shiny.ens-paris-saclay.fr/app/gallicagram}.} Dataset containing text retrieved through OCR.

\subsubsection{GallicaPress}
\uline{Source}: Corpus contributed by OpenLLM partners. A version is also published here: \url{https://huggingface.co/datasets/PleIAs/French-PD-Newspapers}. License: Public domain.

GallicaPress is a large collection of French newspapers and periodicals in the public domain made available through the French National Library (\url{https://gallica.bnf.fr/accueil/fr/content/accueil-fr?mode=desktop}) and extracted from Gallicagram.\footnote{\url{https://shiny.ens-paris-saclay.fr/app/gallicagram}.} Dataset containing text retrieved through OCR.

\subsubsection{Gutenberg}
The Gutenberg Project (\url{https://www.gutenberg.org/}) proposes a collection of free eBooks, manually prepared by human annotators. Our data is extracted from \url{https://github.com/pgcorpus/gutenberg} (CC BY 4.0) and directly from the Project Gutenberg site at \url{http://aleph.gutenberg.org/} (Open license).

\subsubsection{HAL}
\uline{Source}: \url{https://huggingface.co/datasets/bigscience-data/roots_fr_hal_archives_ouvertes}.\\ License: Roots dataset.

The HAL dataset~\citep{roots} is a collection of scientific papers and manuscripts distributed through the open science platform HAL (\url{https://hal.science/}). Dataset containing text retrieved through OCR.

\subsubsection{InterventionsParlement}

InterventionsParlement is a collection of transcripts of speeches made during French parliamentary debates. The data is extracted from Regards Citoyens.\footnote{\url{https://www.regardscitoyens.org/\#&panel1-4}; CC BY-SA.}

\subsubsection{LEGI and OpenData}
\uline{Source}: Corpus contributed by OpenLLM partners. A version is also published at \url{https://huggingface.co/datasets/Nicolas-BZRD/DILA_OPENDATA_FR_2023}.

The OpenData (DILA) dataset\footnote{\url{https://echanges.dila.gouv.fr/OPENDATA/}; Data collection date: October, 2023.} is a French government collection of ``text data extracted from various sources provided by the French government, specifically the Direction de l'Information Légale et Administrative (DILA). This dataset contains a wide range of legal, administrative, and legislative documents. The data has been organized into several categories for easy access and analysis'' (the data card).

\subsubsection{MathPile (Commercial)}
\uline{Source}: \url{https://huggingface.co/datasets/GAIR/MathPile_Commercial}. License: CC BY-SA 4.0.

Mathpile~\citep{mathpile} is a preprocessed collection of documents focused on math, including Textbooks, arXiv, Wikipedia, ProofWiki, StackExchange, and web pages from Common Crawl. The content targets a range of levels, from kindergarten through postgraduate level. MathPile Commercial was obtained by removing documents from MathPile\footnote{\url{https://huggingface.co/datasets/GAIR/MathPile}; CC BY-SA-NC 4.0.} that do not allow commercial use.

\subsubsection{OpenEdition}
OpenEdition\footnote{\url{https://www.openedition.org/}: Open Edition license: \url{https://www.openedition.org/12554}.} is a collection of scientific books, journal articles, blog entries and event descriptions.

\subsubsection{PeS2o (v2)}
\uline{Source}: \url{https://huggingface.co/datasets/allenai/peS2o} version v2. License: ODC BY-v1.0.\footnote{\url{https://github.com/allenai/s2orc/}}

PeS2o~\citep{peS2o} is a preprocessed collection of academic papers designed for pretraining of language models.
PeS2o is composed of two subsets: one containing full papers and one containing only paper titles and abstracts. Dataset containing (some) text retrieved through OCR. Knowledge cutoff: 2023-01-03. PeS2o is extracted from \url{https://github.com/allenai/s2orc}.\footnote{S2ORC; cf.\url{https://aclanthology.org/2020.acl-main.447/}; ODC BY-v1.0.}

\subsubsection{Pile (Uncopyrighted)}
\uline{Source}: \url{https://huggingface.co/datasets/monology/pile-uncopyrighted}. License: Other.\footnote{\url{https://huggingface.co/datasets/monology/pile-uncopyrighted}}

The Pile~\citep{pile,pile2}\footnote{\url{https://huggingface.co/datasets/EleutherAI/pile}; MIT License.} is a collection of various subsets. We selected the following for training Lucie-7B (descriptions taken from~\cite{pile2}):
\begin{itemize}
    \item FreeLaw (\url{https://free.law/}): ``The Free Law Project is US registered non-profit that provide access to millions of legal opinions and analytical tools for academic studies in the legal realm.''
    \item StackExchange (\url{https://stackexchange.com/}): ``The StackExchange dataset is a dump of anonymized user-contributed content on the Stack Exchange network, a popular collection of websites centered around user-contributed questions and answers.''
    \item USPTO Backgrounds (\url{https://bulkdata.uspto.gov/}): ``The USPTO Backgrounds dataset is a set of background sections from patents granted by the United States Patent and Trademark Office, derived from its published bulk archives.''
    \item DM Mathematics (\url{https://github.com/google-deepmind/mathematics_dataset}): ``The DeepMind Mathematics dataset consists of a collection of mathematical problems such as algebra, arithmetic, calculus, number theory, and probability, formatted as natural language prompts~\cite{pile_cited}.''
    \item Ubuntu IRC (\url{https://irclogs.ubuntu.com/}): ``The Ubuntu IRC dataset is derived from the publicly available chatlogs of all Ubuntu-related channels on the Freenode IRC chat server.''
    \item PhilPapers (\url{https://philpapers.org/}): a ``dataset of open access philosophy publications from an international database maintained by the Center for Digital Philosophy at the University of Western Ontario.''
    \item NIH ExPORTER: ``The NIH Grant abstracts provides a bulk-data repository for awarded applications through the ExPORTER service covering the fiscal years 1985-present.''
\end{itemize}

\subsubsection{QuestionsEcritesParlement}
QuestionsEcritesParlement is a collection of long written questions, read during a session at the French National Assembly. Questions are asked by a member of the French parliament and addressed to a minister (who is given two months to respond). Data is extracted from Regards Citoyens.\footnote{\url{https://www.regardscitoyens.org/\#&panel1-4}; CC BY-SA.}

\subsubsection{RedPajama (v2)}
\uline{Source}:
    \url{https://huggingface.co/datasets/togethercomputer/RedPajama-Data-V2}.
    Data license: not specified (data)
    but see Common Crawl terms of use.\footnote{\url{https://commoncrawl.org/terms-of-use}} License for data preparation code: Apache 2.0.
    
\begin{figure}[H]
    \centering
    \includegraphics[width=0.4\linewidth]{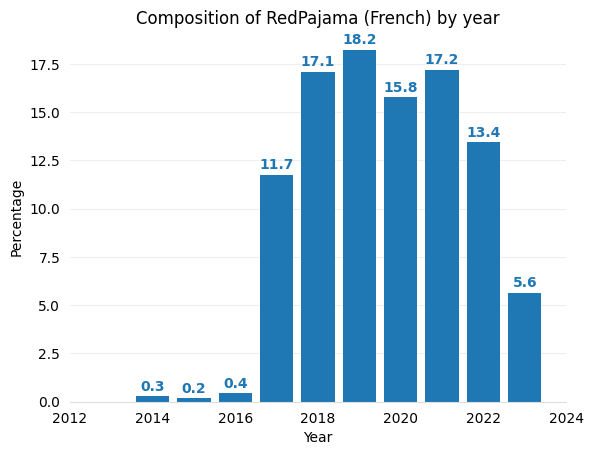}
    \hfill
    \includegraphics[width=0.4\linewidth]{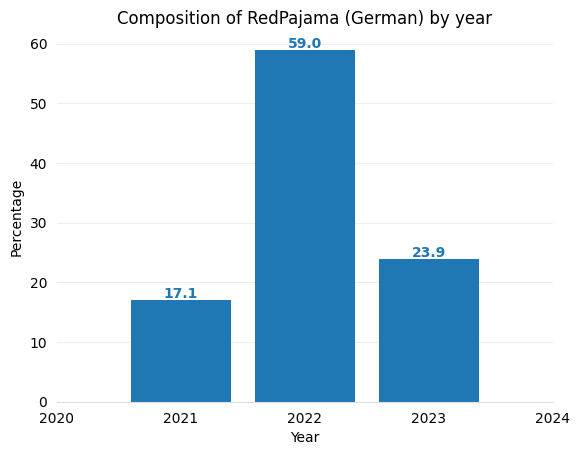}
    \\
    \includegraphics[width=0.4\linewidth]{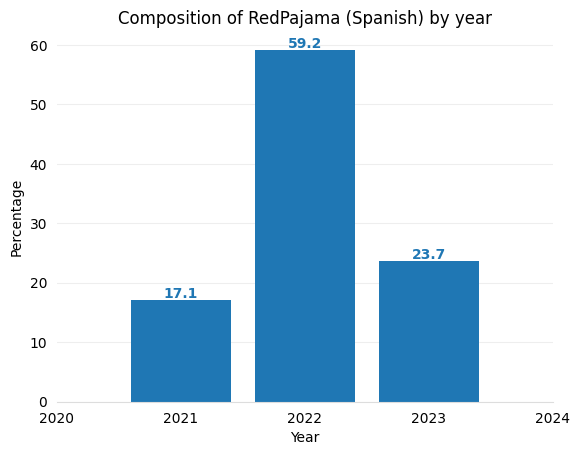}
    \hfill
    \includegraphics[width=0.4\linewidth]{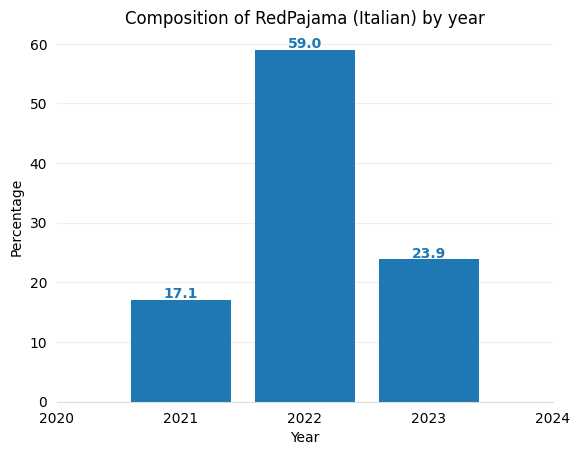}
    \caption{
        Distribution of documents by year in the RedPajama v2 dataset for French, German, Spanish, and Italian.
    }
    \label{fig:distribution_redpajama}
\end{figure}

RedPajama v2~\citep{redpajama} is an ``open dataset for training large language models. The dataset includes over 100B text documents coming from 84 CommonCrawl (\url{https://commoncrawl.org/}) snapshots and processed using the CCNet~\citep{wenzek-etal-2020-ccnet}\footnote{\url{https://github.com/facebookresearch/cc_net}} pipeline. Out of these, there are 30B documents in the corpus that additionally come with quality signals, and 20B documents that are deduplicated'' (from \url{https://github.com/togethercomputer/RedPajama-Data}). Most recent crawl for French data in the Lucie Training Dataset v1.1 (see Section \ref{sec:data-preparation}): 2023-14. Figure~\ref{fig:distribution_redpajama} shows the distribution of documents by year in the RedPajama dataset.

\subsubsection{STAC}
\uline{Source}: \url{https://www.irit.fr/STAC/corpus.html}. License: CC BY-SA-NC 4.0.

STAC~\citep{stac} is a collection of multiparty chats from an online version of the game Settlers of Catan.
The full STAC corpus contains annotations for discourse structure. We use only the text of the chats.

\subsubsection{TheStack (deduplicated v1.2)}
\uline{Source}: \url{https://huggingface.co/datasets/bigcode/the-stack-dedup}. License: Other (mixture of copyleft licenses)\footnote{\url{https://huggingface.co/datasets/bigcode/the-stack-dedup}}.

The Stack~\citep{thestack} ``contains over 6TB of permissively-licensed source code files covering 358 programming languages. The dataset was created as part of the \href{https://www.bigcode-project.org/}{BigCode Project}, an open scientific collaboration working on the responsible development of Large Language Models for Code (Code LLMs). The Stack serves as a pretraining dataset for Code LLMs, i.e., code-generating AI systems which enable the synthesis of programs from natural language descriptions as well as other from code snippets. This is the near-deduplicated version with 3TB data'' (from the Hugging Face data card). Data is extracted from \url{https://github.com/} via GHarchive.\footnote{\url{https://www.gharchive.org/}; mixed licenses for source.}

\subsubsection{Theses}
The theses dataset is a collection of doctoral theses published in France via \url{https://theses.fr/?domaine=theses}\footnote{Licence Ouverte/Open License version 2.0:\\ \url{https://www.data.gouv.fr/fr/datasets/theses-soutenues-en-france-depuis-1985/}} and the open-access platform, HAL.\footnote{\url{https://about.hal.science/}} Dataset containing text retrieved through OCR.

\subsubsection{Wikipedia, Wikisource, Wiktionary}

The Wiki datasets are extracted from the Wikimedia dumps\footnote{\url{https://dumps.wikimedia.org/other/enterprise_html/runs/}.\\ GFDL/CC BY-SA: \url{https://dumps.wikimedia.org/legal.html}} and formatted in markdown. They are also published as separate collections on Hugging Face.\footnote{\url{https://huggingface.co/datasets/OpenLLM-France/wikipedia},\\ \url{https://huggingface.co/datasets/OpenLLM-France/wikisource},\\ \url{https://huggingface.co/datasets/OpenLLM-France/wiktionary}}

These datasets are a plain text version of wikipedia.org  pages for several languages (English, German, French, Spanish, Italian), as well as
\url{https://fr.wikisource.org}
and \url{https://fr.wiktionary.org}.
The Wikipedia datasets were generated from the dump made on February 1, 2024.
The dumps used for fr.wikisource.org and fr.wiktionary.org were those from  December 1, 2023 and December 20, 2023, respectively.

\subsubsection{YouTube}
The YouTube dataset consists of subtitles from YouTube videos collected from \url{https://www.youtube.com/}
by LINAGORA Labs and LeVoiceLab\footnote{\url{https://www.levoicelab.org/}} at the end of 2022.

\subsection{Dataset composition}\label{sec:composition}
\begin{figure}[H]
    \begin{minipage}{\textwidth}
        \centering
        \includegraphics[width=\textwidth]{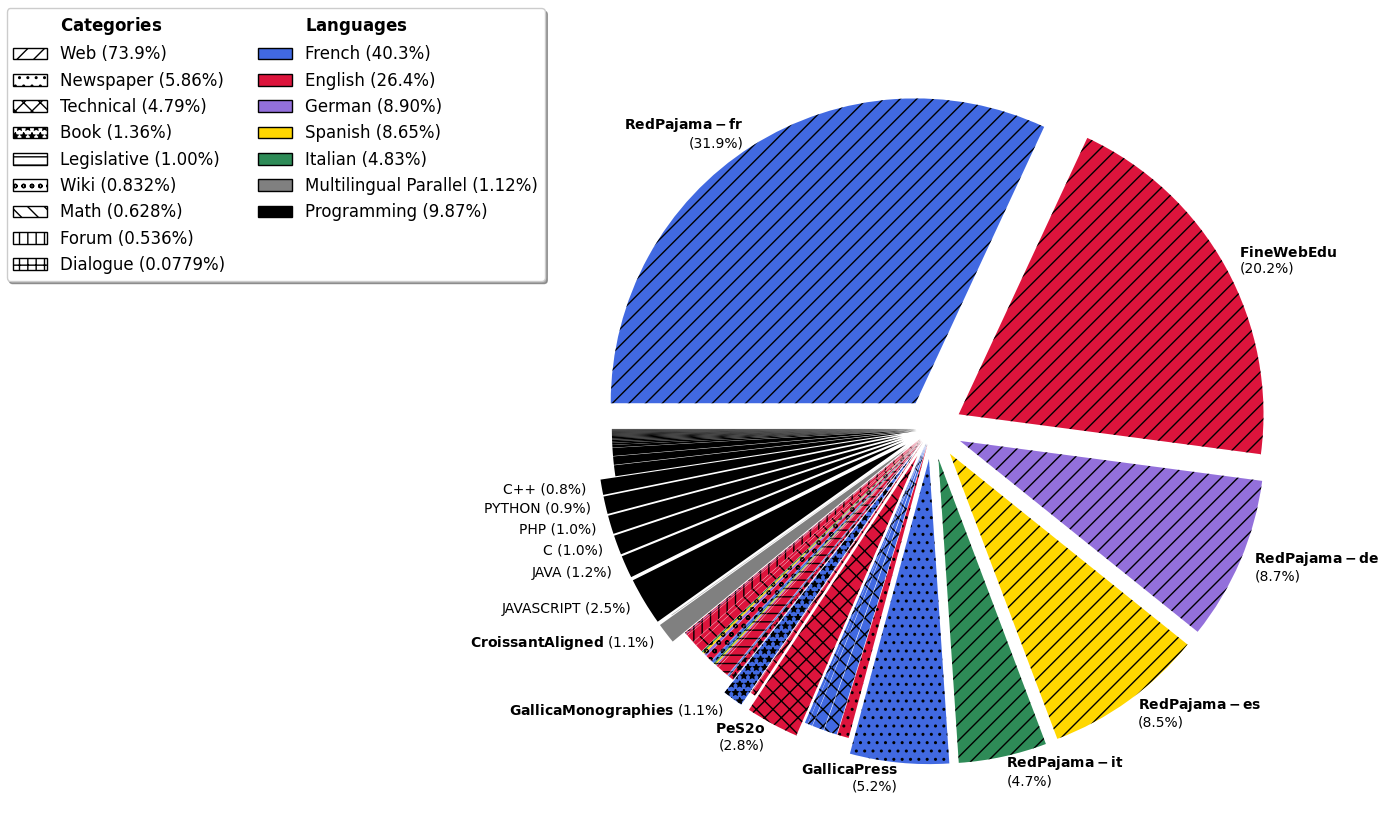}
        \caption{
            Composition of the raw training data (2.32 billion tokens),
            by language (colors) and category (hatch patterns).
        }
        \label{fig:pie_dataset_composition}
    \end{minipage}
\end{figure}

Figure~\ref{fig:pie_dataset_composition}
shows the distribution of the Lucie Training Dataset. Nearly a third of the raw data comes from the French subset of RedPajama, while a fifth of the data comes from FineWebEdu, a subset of FineWeb that has been classified as containing high-quality educational content in English. The remaining data comes from a variety of sources as shown in the ``Categories'' column of the legend of Figure~\ref{fig:pie_dataset_composition}. Overall, 40\% of the data comes from French sources.

\begin{figure}[H]
    \begin{minipage}{\textwidth}
        \centering
        \includegraphics[width=\textwidth]{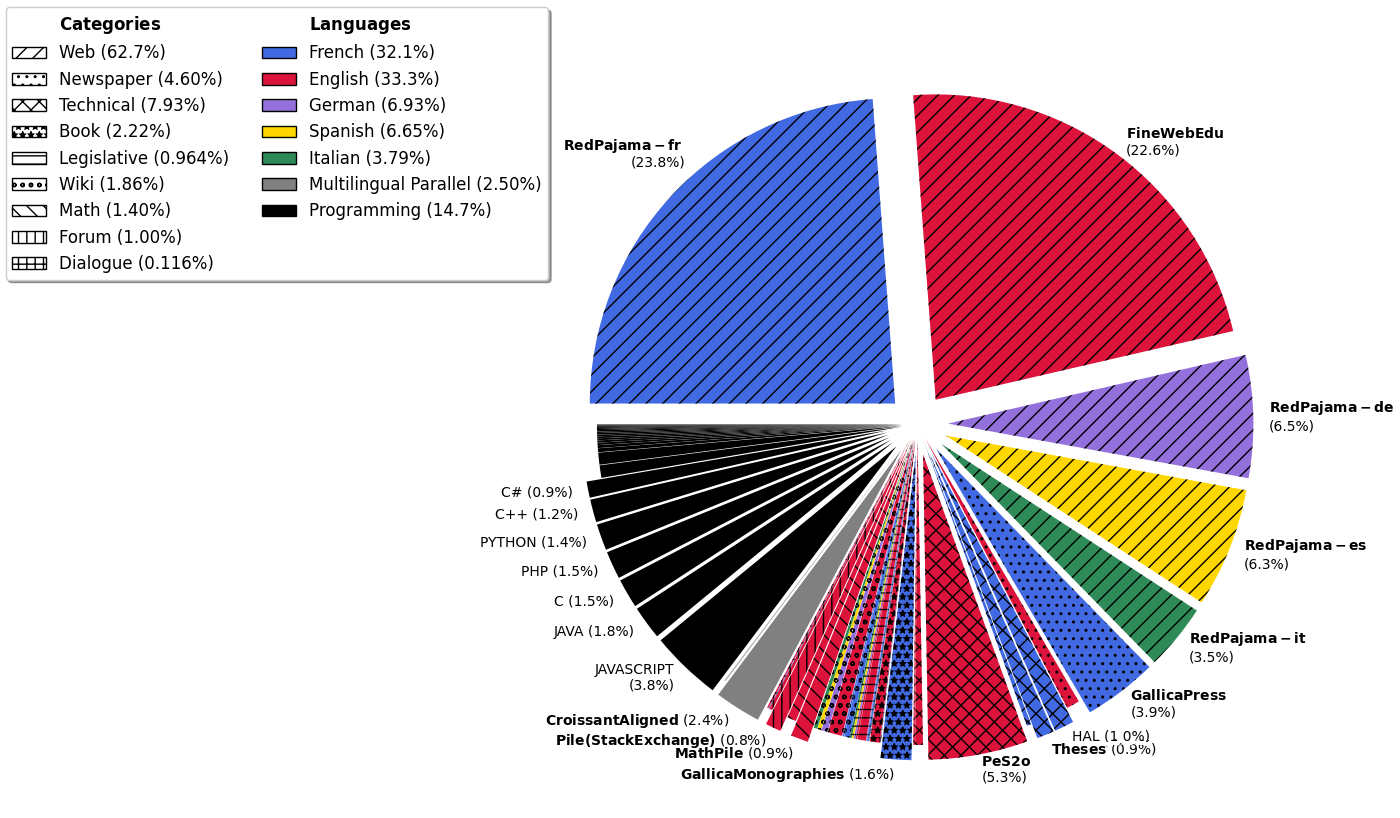}
        \caption{
            Composition of the training data after up-sampling some subsets,
            used in the first phase of Lucie-7B pretraining (3.11 billion tokens).
        }
        \label{fig:pie_dataset_composition_training}
    \end{minipage}
\end{figure}
Figure~\ref{fig:pie_dataset_composition_training} shows the distribution of data that was actually seen by the model after upsampling certain datasets. Higher quality datasets, such as Wikipedia and Gutenberg books, were seen 3 times. Lower-quality data coming from OCR sources such as HAL or Gallica Press, were seen only 1 or 2 times. Very large datasets such as RedPajama and FineWebEdu were only seen once. Details are provided in Table \ref{tab:epochs_datasets}. The resulting dataset composition contains nearly equal amounts of French and English data and a lower overall proportion of French web data.


\begin{table}[h]
    \caption{Epochs per dataset}\label{tab:epochs_datasets}
    \centering
    \begin{tabular}{ll}
    \toprule
    \multicolumn{1}{l}{\textbf{Datasets}} & \multicolumn{1}{r}{\textbf{Epochs}} \\
    \midrule
    Eurovoc (en), GallicaPress, Pile (FreeLaw), RedPajama & 1 \\
    AmericanStories, FineWebEdu & 1.5 \\
    AmendementsParlement, Claire, DiscoursPublics, Europarl, Eurovoc (es/it/de), & 2\\ 
    Gallica Monographies, Gutenberg (fr), HAL, InterventionsParlement, OpenData/LEGI, & 2\\
    Persée, OpenEdition, QuestionsEcritesParlement, STAC, TheStack, Theses, YouTube  &  2\\
    Pile (NIH, PhilPapers, StackExchange, Ubuntu\_IRC, USPTO\_Backgrounds), PeS2o & 2.5 \\
    Gutenberg (de/en/es/it), Pile (DM\_Math), Wikimedia, CroissantAligned, EuroparlAligned & 3\\
    \bottomrule
    \end{tabular}
\end{table}

Table~\ref{tab:datasets_by_language} gives the number of documents, words, tokens, and characters for each language in the dataset.
Table~\ref{tab:dataset_composition} provides an overview of the dataset composition,
broken down by sources and language and grouped by category.
Table~\ref{tab:dataset_composition_programming} shows the distribution of all programming languages in the dataset (TheStack subset).

\begin{table}[H]
    \caption{Composition of the Lucie Training Dataset, broken down by language.}
    \label{tab:datasets_by_language}
    \centering
    \begin{tabular}{lrrrr}
    \toprule
        \textbf{Language} & \textbf{M docs} & \textbf{B words} & \textbf{B tokens} & \textbf{B chars} \\
    \midrule
    fr & 653.812 & 583.687 & 928.618 & 3619.672 \\
    en & 554.289 & 412.202 & 611.894 & 2553.541 \\
    code & 125.769 & 51.306 & 228.954 & 630.749 \\
    de & 165.915 & 105.609 & 206.610 & 764.779 \\
    es & 171.651 & 123.857 & 200.825 & 759.457 \\
    it & 99.440 & 62.051 & 112.031 & 404.454 \\
    fr-en & 410.032 & 17.016 & 25.494 & 107.658 \\
    it-en & 1.901 & 0.100 & 0.151 & 0.638 \\
    es-en & 1.961 & 0.103 & 0.143 & 0.631 \\
    de-fr & 1.792 & 0.0908 & 0.141 & 0.621 \\
    \midrule
    \textbf{TOTAL} & 2186.562 & 1356.021 & 2314.862 & 8842.200 \\
    \bottomrule
    \end{tabular}
\end{table}

\begin{longtable}{llcrrrr}
    \caption{
        Composition of the Lucie Training Dataset, broken down by category, language and original source.
        All statistics in this table are also available at \url{https://huggingface.co/datasets/OpenLLM-France/Lucie-Training-Dataset/blob/main/metadata/dataset_composition.csv}.
        Token counts are computed using the tokenizer for Lucie-7B.
    }
    \label{tab:dataset_composition}
    \\
    \toprule
    \textbf{Category} & \textbf{Dataset} & \textbf{Language} & \textbf{M docs} & \textbf{B words} & \textbf{B tokens} & \textbf{B chars} \\
    \endfirsthead
    \endhead
    \midrule
        \textbf{Web} & RedPajama & fr & 640.770 & 477.758 & 741.023 & 2974.596 \\
        & FineWebEdu & en & 421.209 & 327.453 & 467.837 & 2018.215 \\
        & RedPajama & de & 162.779 & 103.078 & 201.371 & 747.631 \\
        &    & es & 169.447 & 121.751 & 197.125 & 746.984 \\
        &    & it & 97.324 & 60.194 & 108.416 & 393.012 \\
    \midrule
        \textbf{Newspaper} & GallicaPress & fr & 3.205 & 67.496 & 121.606 & 408.882 \\
        & AmericanStories & en & 59.420 & 8.902 & 14.313 & 50.844 \\
    \midrule
        \textbf{Technical} &
        PeS2o & en & 38.972 & 42.296 & 65.365 & 268.963 \\
        &HAL & fr & 0.349 & 9.356 & 16.224 & 58.308 \\
        &Theses & fr & 0.102 & 7.547 & 14.060 & 47.758 \\
        &Pile (USPTO\_Backgrounds) & en & 5.139 & 3.492 & 5.105 & 22.309 \\
        &OpenEdition & fr & 0.939 & 2.225 & 3.604 & 14.459 \\
        &Pile (PhilPapers) & en & 0.0308 & 0.363 & 0.618 & 2.304 \\
        &Pile (NIH\_ExPorter) & en & 0.914 & 0.288 & 0.431 & 1.979 \\
    \midrule
        \textbf{Book} & 
        GallicaMonographies & fr & 0.278 & 15.106 & 25.169 & 90.456 \\
        & Gutenberg & en & 0.0563 & 3.544 & 5.516 & 20.579 \\
        &    & fr & 0.00345 & 0.227 & 0.383 & 1.392 \\
        &    & de & 0.00188 & 0.0987 & 0.193 & 0.654 \\
        &    & it & 0.000958 & 0.0657 & 0.129 & 0.414 \\
        &    & es & 0.000735 & 0.0512 & 0.0920 & 0.303 \\
    \midrule
        \textbf{Legislative Texts} & 
        Pile (FreeLaw) & en & 3.415 & 8.204 & 14.011 & 52.580 \\
        & Eurovoc & en & 0.272 & 1.523 & 2.571 & 9.468 \\
        &    & it & 0.245 & 0.731 & 1.527 & 4.867 \\
        &    & de & 0.247 & 0.678 & 1.497 & 4.915 \\
        &    & es & 0.246 & 0.757 & 1.411 & 4.684 \\
        & OpenData & fr & 1.169 & 0.755 & 1.209 & 4.638 \\
        & QuestionsEcritesParlement & fr & 0.189 & 0.108 & 0.156 & 0.705 \\
        & LEGI & fr & 0.621 & 0.0878 & 0.145 & 0.563 \\
        & AmendementsParlement & fr & 0.673 & 0.0452 & 0.0738 & 0.274 \\
    \midrule
        \textbf{Legislative Transcripts} &
        DiscoursPublics & fr & 0.110 & 0.163 & 0.238 & 1.025 \\
        &InterventionsParlement & fr & 1.832 & 0.104 & 0.157 & 0.654 \\
        &Europarl & de & 0.0102 & 0.0451 & 0.0734 & 0.327 \\
        &    & es & 0.0103 & 0.0524 & 0.0733 & 0.325 \\
        &    & fr & 0.0103 & 0.0528 & 0.0717 & 0.339 \\
        &    & en & 0.0111 & 0.0563 & 0.0690 & 0.339 \\
    \midrule
        \textbf{Wiki} &
        Wikipedia & en & 6.893 & 4.708 & 7.898 & 26.616 \\
        &    & de & 2.877 & 1.709 & 3.476 & 11.252 \\
        &    & fr & 2.648 & 1.726 & 2.940 & 9.879 \\
        &    & es & 1.947 & 1.245 & 2.124 & 7.161 \\
        &    & it & 1.870 & 1.060 & 1.959 & 6.161 \\
        &Wikisource & fr & 0.186 & 0.523 & 0.795 & 3.080 \\
        &Wiktionary & fr & 0.650 & 0.0531 & 0.117 & 0.347 \\
    \midrule
        \textbf{Math} &
        MathPile & en & 0.737 & 3.408 & 9.637 & 27.290 \\
        &Pile (DM\_Mathematics) & en & 0.992 & 1.746 & 4.928 & 8.127 \\
    \midrule
        \textbf{Forum} &
        Pile (StackExchange) & en & 15.269 & 4.534 & 10.275 & 33.609 \\
        &Pile (Ubuntu\_IRC) & en & 0.0104 & 0.867 & 2.159 & 5.610 \\
    \midrule
        \textbf{Dialogue} &
        Claire & en & 0.949 & 0.818 & 1.161 & 4.709 \\
        &        & fr & 0.0393 & 0.210 & 0.311 & 1.314 \\
        &YouTube & fr & 0.0375 & 0.145 & 0.336 & 1.003 \\
        &Stac & en & 0.0000450 & 0.0000529 & 0.000121 & 0.000327 \\
    \midrule
        \textbf{Multilingual Parallel} &
        CroissantAligned & fr-en & 408.029 & 16.911 & 25.351 & 107.003 \\
        &EuroparlAligned & it-en & 1.901 & 0.100 & 0.151 & 0.638 \\
        &    & fr-en & 2.003 & 0.105 & 0.143 & 0.655 \\
        &    & es-en & 1.961 & 0.103 & 0.143 & 0.631 \\
        &    & de-fr & 1.792 & 0.0908 & 0.141 & 0.621 \\
    \midrule
        \textbf{Programming} &
        TheStack & code & 125.769 & 51.306 & 228.954 & 630.749 \\
    \bottomrule
\end{longtable}

\begin{table}[t]
    \caption{
    Distribution of programming languages included in the Lucie Training Dataset. Data taken from TheStack.
    All numbers in this table are available in the CSV file at \url{https://huggingface.co/datasets/OpenLLM-France/Lucie-Training-Dataset/blob/main/metadata/dataset_composition.csv}.
    Token counts are computed using the tokenizer for Lucie-7B.
        }
    \label{tab:dataset_composition_programming}

    \centering
    \begin{tabular}{lrrrr}
    \toprule
        \textbf{Language} & \textbf{M docs} & \textbf{B words} & \textbf{B tokens} & \textbf{B chars} \\
    \midrule
    JAVASCRIPT & 21.109 & 8.526 & 58.609 & 141.647 \\
    JAVA & 20.152 & 7.421 & 27.680 & 89.297 \\
    C & 8.626 & 5.916 & 24.092 & 57.428 \\
    PHP & 15.905 & 4.865 & 22.883 & 66.844 \\
    PYTHON & 12.962 & 5.434 & 21.683 & 64.304 \\
    C++ & 6.378 & 4.584 & 18.835 & 50.892 \\
    C\# & 10.839 & 3.574 & 13.381 & 46.286 \\
    GO & 4.730 & 2.735 & 10.262 & 25.738 \\
    TYPESCRIPT & 10.637 & 2.617 & 9.836 & 28.815 \\
    RUST & 1.387 & 0.872 & 3.241 & 9.529 \\
    RUBY & 3.405 & 0.646 & 2.392 & 7.139 \\
    SWIFT & 1.756 & 0.553 & 1.876 & 6.134 \\
    KOTLIN & 2.243 & 0.454 & 1.758 & 5.769 \\
    SCALA & 1.362 & 0.457 & 1.587 & 4.862 \\
    TEX & 0.398 & 0.394 & 1.507 & 3.805 \\
    LUA & 0.559 & 0.318 & 1.367 & 3.279 \\
    DART & 0.933 & 0.308 & 1.242 & 3.864 \\
    PERL & 0.392 & 0.297 & 1.149 & 2.634 \\
    MATHEMATICA & 0.0269 & 0.120 & 1.117 & 1.720 \\
    ASSEMBLY & 0.248 & 0.209 & 0.867 & 1.575 \\
    HASKELL & 0.545 & 0.307 & 0.807 & 2.364 \\
    FORTRAN & 0.165 & 0.192 & 0.780 & 1.843 \\
    JULIA & 0.299 & 0.152 & 0.660 & 1.539 \\
    OCAML & 0.160 & 0.130 & 0.430 & 1.107 \\
    ERLANG & 0.0994 & 0.0657 & 0.260 & 0.726 \\
    ELIXIR & 0.282 & 0.0731 & 0.258 & 0.737 \\
    CLOJURE & 0.126 & 0.0448 & 0.179 & 0.492 \\
    R & 0.0392 & 0.0278 & 0.158 & 0.305 \\
    MATLAB & 0.000967 & 0.00865 & 0.0427 & 0.0372 \\
    RACKET & 0.00420 & 0.00479 & 0.0153 & 0.0378 \\
    \bottomrule
    \end{tabular}
\end{table}

\subsection{Data preparation}\label{sec:data-preparation}
In this section we describe the different pre-processing methods applied to improve the quality of our training dataset and to abide by our principles regarding data rights and privacy. In particular, we discuss how we filtered documents processed with OCR, selected web data, formatted parallel translation data and standardized metadata, among other filtering and cleaning methods.

We note that pre-processing was done in two stages leading to two versions of the Lucie Training Dataset, v1.1 and v1.2. The latter was produced during the context length extension phase described in Section \ref{sec:pretraining2} and targeted four datasets used in this phase: Gallica Monographies, Gallica Press, and Ubuntu\_IRC and PhilPapers from the Pile.

\subsubsection{OCR data filtering}

In an effort to remove particularly noisy OCR documents, we evaluated some of our datasets for perplexity. The use of small language models for corpus perplexity evaluation and data selection was introduced in the CCNet paper~\citep{wenzek-etal-2020-ccnet}, where 5-gram Kneser-Ney language models were trained for 48 languages using their respective Wikipedia articles. These models are highly advantageous due to their low training and inference costs, especially when implemented with the KenLM~\citep{heafield-2011-kenlm} library. They are trained on text tokenized with SentencePiece, which segments text into subwords based on statistical patterns, improving the models' ability to estimate perplexity for unusual character sequences. We note  that RedPajama also uses CCNet models to filter CommonCrawl content as part of its metrics aggregation. 

We find in practice that these models provide effective signals for filtering problematic OCR texts, and we manually determined for each corpus the ideal perplexity threshold below which we did not observe a high incidence of OCR problems. Our code for computing CCNet perplexity with parallelization on arbitrary parquet files is available at \url{https://github.com/OpenLLM-France/Lucie-dataset-filtering}. Table \ref{tab:perplexity} lists the thresholds chosen per OCR-based corpus (where documents over the threshold are discarded):

\begin{table}[H]
    \caption{CCNet perplexity scores chosen as thresholds for discarding particularly noisy OCR documents.}
    \label{tab:perplexity}
    \centering
    \begin{tabular}{lr}
    \toprule
        \textbf{Corpus} & \textbf{CCNet threshold} \\
    \midrule
        American Stories & 2310 \\
        Eurovoc & 1500 \\
        HAL & 930 \\
        Thèses & 2000 \\
    \bottomrule
    \end{tabular}
\end{table}

The process was slightly more elaborate for Gallica Monographies and Gallica Press. In version 1.1 of the dataset, documents from these corpora were split into chunks and chunks were kept if the source language was detected as French by \texttt{FastText}~\citep{fasttext} with a confidence score of 0.65 or above, and the perplexity score, as measured using a CCNET model in French, was between 10 and 1000. In version 1.2, we used OCR scores already available in the metadata of the source corpora. In this case, documents were scored on a scale of 0-100, with 100 being the highest quality. We kept documents with an OCR score of at least 90 out of 100.

\subsubsection{Web data filtering}

For both FineWebEdu and RedPajama, we filtered out urls whose base domain overlapped with a dataset already in the Lucie Training Dataset (e.g., \texttt{philpapers.org}, \texttt{theses.fr}) in order to increase diversity of content. In an effort to select data free from opt-out evidence according to the 4th article of the copyright European directive (2019), we also filtered out any urls that either did not have a \texttt{robots.txt} or that disallowed the CCBot as of July 2024.

For RedPajama, we performed further filtering and deduplication in an effort to select higher quality data. Among the metadata in RedPajama are tags indicating toxicity of certain urls based on the blacklists in \url{https://dsi.ut-capitole.fr/blacklists/}. We filtered out urls from this blacklist.

A series of filters based on quality signals available in RedPajama was also used.\footnote{\url{https://github.com/togethercomputer/RedPajama-Data?tab=readme-ov-file\#quality-annotations}} In particular, we used CCNet perplexity to filter out documents with perplexity below 10 or above 1000. We used C4 filtering, which included removal of documents containing toxic words. We also used gopher filtering and repetition removal as document deduplication. Finally, we used regular expressions to replace email addresses and IP addresses with random addresses and used the \texttt{Datatrove}\footnote{\url{https://github.com/huggingface/datatrove}} library to perform MinHash deduplication on each snapshot and language independently, following practices in FineWeb. (\texttt{Datatrove} was also used to manage the other url filtering described above, though filtering was based on scores provided by RedPajama rather than scores recalculated with \texttt{Datatrove}.)

\subsubsection{Extraction of YouTube videos}

The extraction pipeline consisted of the following steps:
\begin{enumerate}
    \item \textbf{Searching for YouTube videos likely in French:}  
    Based on searches generated automatically from random sequences of words extracted from a corpus of French journalistic articles (initially obtained through a web crawling tool applied to publicly accessible news and media sites such as Huffington Post, 20 Minutes, Le Parisien, Actu, Numerama, Slate, {\it etc.}).  
    Selection of videos with subtitles labeled as ``French,'' excluding those marked as ``automatically generated.''  
    \\ \textit{At this stage: 52,778 videos selected, corresponding to 10,654 hours of audio.}

    \item \textbf{Selection of videos whose subtitle language classification confirms French with a certain confidence score}  
    \\ \textit{At this stage: 51,934 videos selected, corresponding to 10,425 hours of audio.}

    \item \textbf{Selection of videos whose subtitles contain uppercase, lowercase, and punctuation marks:}  
    This step filters out automatically generated subtitles created with speech recognition tools.  
    \\ \textit{At this stage: 45,488 videos selected, corresponding to 8,904 hours of audio.}

    \item \textbf{Extraction of audio tracks from the selected videos.}

    \item \textbf{Automatic formatting of transcripts obtained from subtitles:}  
    Removal of emojis, sound event annotations in brackets (like ``[Music]'') and extra text such as ``subtitled by XXX'' (shown during the final seconds of the video).\footnote{Removing subtitle meta-annotations was done to avoid an unwanted bias we observed when performing Automatic Speech Recognition with OpenAI's Whisper model, trained on a similar Youtube crawling dataset: the model tends to predict things like ``subtitles from amara.org community'' on input audio with just silence.}

    \item \textbf{Selection of videos where an automatic speech recognition tool correctly transcribes the first 30 seconds with a minimum recall and precision rate:}  
    \\ \textit{At this stage: 37,513 videos selected, corresponding to 7,541 hours of audio.}

    \item \textbf{Realignment of the transcript:}  
    Ensuring accurate timestamps in the transcriptions based on the subtitles and excluding audios where alignment fails.  
    \\ \textit{At this stage: 36,618 videos selected, corresponding to 6,729 hours of audio, including 4,293 hours of speech (after removing silence and music backgrounds).}
\end{enumerate}

\subsubsection{Multilingual aligned corpora}

The original text field of the CroissantLLM dataset\footnote{\url{https://huggingface.co/datasets/croissantllm/croissant_dataset_no_web_data}} contains a sentence or passage in French or English immediately followed by its translation without any indication of which passage is in which language. To prepare these data for training, we first  split each text into separate, monolingual passages and then tagged each passage with the appropriate language code, identified automatically using the \texttt{langid} Python library.\footnote{\url{https://pypi.org/project/langid/}} In the Lucie Training Dataset, the \texttt{extra} metadata field for CroissantAligned contains separate keys, \texttt{text\_fr} for French and \texttt{text\_en} for English, that store the texts separately (see Section \ref{sec:metadata}). The Europarl parallel data were already tagged by language. The \texttt{extra} field in the metadata for EuroparlAligned provides texts in the two languages under the sub-fields \texttt{text\_1} and \texttt{text\_2}, and the corresponding language codes under \texttt{lang\_1} and \texttt{lang\_2}.

For both Croissant Aligned and Europarl parallel data, each monolingual text was paired with its translation, but random separators and various methods of prefixing the text with the language (name or code) were added. This was done as a precaution to prevent models trained on this data from switching languages when generating text and can be seen as a very basic instruction to translate the source (first) text into the target (second) text.

\subsubsection{Formatting of Wiki datasets}

The text for the Wiki datasets is plain, without HTML tags or wiki templates, but includes markdown syntax for headers, lists and tables,
and some formulas in \LaTeX. The code used to convert Wikimedia dumps to the dataset is available under a GPLv3 license at \url{https://github.com/OpenLLM-France/wikiplaintext}.
Some notes about text formatting:
\begin{itemize}
    \item Superscripts and subscripts are kept as unicode characters when possible and when not in math formulas. For instance: $\mbox{XIII}^{\mbox{e}}$~siècle, 3\,000~m², $\mbox{P}_{\mbox{ 2 }}\mbox{O}_{\mbox{ 7}}~^{\mbox{4-}}$.
    \item Tables are formatted in markdown. For instance:

\begin{minipage}[b]{0.8\textwidth}
\centering
{\small
\begin{verbatim}
| Français | Espéranto | IPA |
| Salut, bonjour | Saluton | [sa.'lu.ton] |
| Oui | Jes | ['jes] |
| Non | Ne | ['ne] |
| Bonsoir | Bonan vesperon | ['bo.nan ves.'pe.ron] |
| Bonne nuit | Bonan nokton | ['bo.nan 'nok.ton] |
\end{verbatim}
}
\end{minipage}

    \item Lists are formatted in markdown. For instance:

\begin{minipage}[b]{0.8\textwidth}
\centering
{\small
\begin{verbatim}
* 1 000 personnes ont l'espéranto comme langue maternelle ;
* 10 000 personnes parlent l'espéranto avec un niveau proche d'une langue maternelle ;
* 100 000 personnes parlent couramment l'espéranto ;
\end{verbatim}
}
\end{minipage}

    \item Headers are formatted in markdown. For instance:

\begin{minipage}[b]{0.8\textwidth}
\centering
{\small
\begin{verbatim}
# Espéranto

L'espéranto est une langue construite internationale utilisée comme langue véhiculaire […]
Fondée sur une grammaire régulière sans exception, l'espéranto est une langue globaleme[…]
C’est en 1887 que Louis-Lazare Zamenhof, sous le pseudonyme Doktoro Esperanto (Docteur […]
L’Association universelle d’espéranto, fondée en 1908, est en relation officielle avec […]

## Définition
### Nom

Le pseudonyme «Doktoro Esperanto» (Docteur « Espérant »), utilisé par Zamenhof pour pré[…]

#### Utilisation du mot espéranto en tant que métaphore

Le nom espéranto fonctionne comme un nom propre quand il désigne la langue même, mais e[…]
Dans le domaine de l'informatique, Java fut qualifié d'« espéranto des langages de prog[…]
\end{verbatim}
}
\end{minipage}

\end{itemize}


\subsubsection{Further filtering and cleaning}

\uline{DiscoursPublics}: the mention of the source URL and the number of views were removed from the text.

\uline{Eurovoc}: Mentions of Credit Institutions Directives (CID) that appear in the raw texts such as ``\texttt{(cid:146)}'' were removed.

\uline{Gutenberg}: Books were filtered based on their author's date of death, so that only texts from authors who died more than 70 years ago are included (80 years for French authors). This filtering was done to ensure that the texts are in the public domain. For each document, headers and and footers containing information about Project Gutenberg were removed.
    
\uline{MathPile (Commercial)}: The content of StackExchange questions and answers were converted to match the \texttt{\{"text": value\}} format described in \ref{sec:metadata} below, using the following python code:
    \begin{verbatim}
    text = sample["question"]["Body"] + "\n\n".join(
        [answer["Body"] for answer in sample["answers"]]
    )
    \end{verbatim}

\uline{Pile (Uncopyrighted)}: In version 1.2 only, papers were removed from PhilPapers if their language, detected using Stanza,\footnote{\url{https://github.com/stanfordnlp/stanza}} was not classified as English, French, German, Spanish or Italian. Texts were removed from Ubuntu IRC if they came from channels predominantly in languages other than English, French, German, Spanish or Italian. Certain encoding errors were also fixed.
    
\uline{Thèses}: Title pages about HAL,\footnote{\url{https://theses.hal.science/}} pages containing a significant fraction of control characters, and duplicate lines were removed. Texts with fewer than 1\,000 words or 10\,000 characters were removed.


\subsubsection{Metadata normalization}\label{sec:metadata}

Each sample in the Lucie Training Dataset comes with the following fields:
\begin{itemize}
    \item ``\texttt{text}'':
        the text content of the document.
    \item ``\texttt{language}'':
        the language of the text sample (note that this information is taken from the original data source and may be incorrect).
        Possible values are:
        \begin{itemize}
            \item the ISO 639-1 code for a given natural language (``\texttt{en}'', ``\texttt{fr}'', ``\texttt{de}'', ``\texttt{es}'', or ``\texttt{it}''),
            \item the name of a programming language prefixed by ``\texttt{code:}\ldots'' (``\texttt{code:python}'', ``\texttt{code:c++}'', \ldots), or
            \item a list of ISO 639-1 codes separated by commas for data containing parallel translations
            (``\texttt{fr,en}'', ``\texttt{de,fr}'', ``\texttt{es,en}'', ``\texttt{it,en}'',
            or one of those pairs in the opposite order if the languages appear in the opposite order in the text).
        \end{itemize}
    \item ``\texttt{source}'':
        an identifier for the source(s) of the text sample.
        Possible values for the source are the following camel-cased strings listed, and correspond to subsets detailed in Section~\ref{sec:datasets}:
        AmendementsParlement,
        AmericanStories,
        Claire,
        CroissantAligned,
        DiscoursPublics,
        Europarl, EuroparlAligned,
        Eurovoc,
        FineWebEdu,
        GallicaMonographies,
        GallicaPress,
        Gutenberg,
        HAL,
        InterventionsParlement,
        LEGI,
        MathPile,
        OpenData,
        OpenEdition,
        PeS2o,
        Pile,
        QuestionsEcritesParlement,
        RedPajama,
        STAC,
        TheStack,
        Theses,
        Wikipedia, Wikisource, Wiktionary,
        YouTube.
    \item ``\texttt{id}'': an identifier that is unique among documents from the same source.
    \item ``\texttt{url}'' (optional): the URL of the original text sample on the web, if available.
    \item ``\texttt{title}'' (optional): the title of the original text sample, if available.
    \item ``\texttt{author}'' (optional): the author of the original text sample, if available.
    \item ``\texttt{date}'' (optional): the publication date of the original text sample, if available.
    \item ``\texttt{quality\_signals}'' (optional):
        a list of quality signals for the text sample in JSON format (which could be used for further filtering or sample weighting).
        It can include indicators computed by `fasttext` and `CCNet`, statistics about occurrences of characters, words, special characters, etc.
    \item ``\texttt{extra}'' (optional):
        extra information about the text sample, in JSON format.
        This can include metadata about the source subset, the rights, etc.
\end{itemize}

Here is an example from the Lucie Training Dataset:
{
    \small
    \begin{verbatim}
    {
        "text": "DROPOSALS FOR THE ERECTION AMI\na completion of Building, in accordance with…",
        "language": "en",
        "source": "AmericanStories",
        "id": "16_1858-12-04_p3_sn84038814_00279557177_1858120401_0390",
        "url": "",
        "title": "miscellaneous.\n\n\nNOTICE\nTO BUILDERS.",
        "author": "",
        "date": "1858-12-04",
        "quality_signals": "{\"char_count\": 670, \"word_count\": 116, \"ccnet_language_score\": 0.939,
                             \"ccnet_perplexity\": 1389.5, \"fasttext_language\": \"en\"}",
        "extra": "{\"newspaper_name\": \"Daily national Democrat\", \"edition\": \"01\", \"page\": \"p3\"}"
    }
    \end{verbatim}
}

\subsection{Dataset release}

The dataset can be found on Hugging Face
at \url{https://huggingface.co/datasets/OpenLLM-France/Lucie-Training-Dataset}.

Configurations are available to load only a subset of the data, e.g.,
\begin{itemize}
    \item only the data in a given language
        (e.g., French, all code data, or just a specific programming language),
    \item only the data from a given source
        (e.g., only the data from the \texttt{RedPajama} source),
    \item only the data from a given source in a given language
        (e.g., only data from \href{https://fr.wikipedia.org/}{wikipedia.fr}),
\end{itemize}

Here are some examples of python code to load the dataset:\\

\hspace{5mm} \begin{minipage}[b]{0.8\textwidth}
\centering
{\small
\begin{minted}[linenos]{python}
from datasets import load_dataset

kwargs = dict(split="train", streaming=True)

dataset_french       = load_dataset("OpenLLM-France/Lucie-Training-Dataset", "fr", **kwargs)

dataset_wikipedia    = load_dataset("OpenLLM-France/Lucie-Training-Dataset", "Wikipedia", **kwargs)

dataset_wikipedia_fr = load_dataset("OpenLLM-France/Lucie-Training-Dataset", "Wikipedia-fr", **kwargs)

dataset_code         = load_dataset("OpenLLM-France/Lucie-Training-Dataset", "code", **kwargs)

dataset_code_python  = load_dataset("OpenLLM-France/Lucie-Training-Dataset", "code-python", **kwargs)
\end{minted}
}
\end{minipage}

\vspace{4mm}
The list of all the possible configurations can be obtained by running the following code:\\

\hspace{5mm} \begin{minipage}[b]{0.8\textwidth}
{\small
\begin{minted}[linenos]{python}
from datasets import load_dataset_builder

config_names = list(load_dataset_builder("OpenLLM-France/Lucie-Training-Dataset").builder_configs)

print(config_names)
\end{minted}
\begin{alltt}
['default', 'en', 'fr', 'de', 'es', 'it', 'de,fr', 'es,en', 'fr,en', 'it,en', 'natural', 'code',
 'code-assembly', 'code-c', 'code-c\#', 'code-c++', 'code-clojure', 'code-dart', 'code-elixir',
 'code-erlang', 'code-fortran', 'code-go', 'code-haskell', 'code-java', 'code-javascript', 
 'code-julia', 'code-kotlin', 'code-lua', 'code-mathematica', 'code-matlab', 'code-ocaml',  
 'code-perl', 'code-php', 'code-python', 'code-r', 'code-racket', 'code-ruby', 'code-rust',  
 'code-scala', 'code-swift', 'code-tex', 'code-typescript', 'AmendementsParlement', 'AmericanStories', 
 'Claire', 'Claire-en', 'Claire-fr', 'CroissantAligned', 'DiscoursPublics', 'Europarl', 'Europarl-de', 
 'Europarl-en', 'Europarl-es', 'Europarl-fr', 'EuroparlAligned', 'EuroparlAligned-de,fr', 
 'EuroparlAligned-es,en', 'EuroparlAligned-fr,en', 'EuroparlAligned-it,en', 'Eurovoc', 'Eurovoc-de', 
 'Eurovoc-en', 'Eurovoc-es', 'Eurovoc-it', 'FineWebEdu', 'GallicaMonographies', 'GallicaPress', 
 'Gutenberg', 'Gutenberg-de', 'Gutenberg-en', 'Gutenberg-es', 'Gutenberg-fr', 'Gutenberg-it', 'HAL', 
 'InterventionsParlement', 'LEGI', 'MathPile', 'OpenData', 'OpenEdition', 'PeS2o', 'PeS2o-s2ag', 
 'PeS2o-s2orc', 'Pile', 'Pile-DM_Mathematics', 'Pile-FreeLaw', 'Pile-NIH_ExPorter', 'Pile-PhilPapers', 
 'Pile-StackExchange', 'Pile-USPTO_Backgrounds', 'Pile-Ubuntu_IRC', 'QuestionsEcritesParlement', 
 'RedPajama', 'RedPajama-de', 'RedPajama-es', 'RedPajama-fr', 'RedPajama-it', 'Stac', 'TheStack', 
 'Theses', 'Wikipedia', 'Wikipedia-de', 'Wikipedia-en', 'Wikipedia-es', 'Wikipedia-fr', 'Wikipedia-it', 
 'Wikisource', 'Wiktionary', 'YouTube']
\end{alltt}
}
\end{minipage}

By default, the dataset on Hugging Face corresponds to the one used in
the first phase of Lucie-7B pretraining (Section \ref{sec:pretraining1}), but other versions are available.
For instance, the following code can be run to load
a more recent version of the data,
where some subsets were improved in terms of quality (Section~\ref{sec:data-preparation}),
and used in the second phase of Lucie-7B pretraining
(context length extension, explained in section~\ref{sec:pretraining2}):\\





\hspace{5mm} \begin{minipage}[b]{0.8\textwidth}
{\small
\begin{minted}[linenos]{python}
from datasets import load_dataset

kwargs = dict(split="train", streaming=True)

name = None # or a configuration (e.g. "fr", "code-python", "Wikipedia-fr", "Pile-PhilPapers")

dataset_v2 = load_dataset("OpenLLM-France/Lucie-Training-Dataset", name, revision="v1.2", **kwargs)
\end{minted}
}
\end{minipage}


\section{Lucie-7B LLM}\label{sec:model}
Lucie-7B is a decoder-only model trained on a causal language modeling task, i.e., it learns to predict the next token. The model was trained on Jean-Zay,\footnote{\url{http://www.idris.fr/eng/jean-zay/jean-zay-presentation-eng.html}} the French national supercomputing cluster, using Megatron-Deepspeed, a library designed for distributed training. This library combines the general GPT architecture and tensor parallelism from Megatron-LM ~\cite{megatronlm1, megatronlm2, megatronlm3} with DeepSpeed's efficient Zero Optimizer, as well as pipeline and data parallelism. The training code is available at \url{https://github.com/OpenLLM-France/Lucie-Training}. It is based on a fork of \texttt{Megatron-DeepSpeed} available at \url{https://github.com/OpenLLM-France/Megatron-DeepSpeed}.

\subsection{Neural network architecture}

Lucie-7B is a transformer with the same neural network architecture as Llama3.1, with the exception of the feed-forward hidden size and vocabulary size.\footnote{\url{https://huggingface.co/meta-llama/Llama-3.1-8B}}
It has exactly 6,706,958,336 free parameters,
with the hyperparameter values listed in Table~\ref{tab:archi_hyperparameters}.
The $\theta$ parameter of Rotary Positional Embedding (RoPE) was 
set to 500,000 during the first phase of pretraining (Section~\ref{sec:pretraining1}) and increased to 20 million during context length extension (Section~\ref{sec:pretraining2}). The higher value was maintained for the rest of the training process.


\begin{table}[H]
\caption{
    Hyperparameters of Lucie-7B's architecture. 
    \usedapache The~$\theta$~parameter of Rotary Positional Embedding (RoPE) was increased during context length extension.
}
\label{tab:archi_hyperparameters}
\centering
\begin{tabular}{lr@{\hspace{0pt}}l}
\midrule
    \textbf{Hyperparameter} & \textbf{Value} \\
    \midrule
    Vocabulary size (\# tokens) & 65,024 \\
    \# transformer blocks & 32 \\
    \# attention heads & 32 \\
    \# key-value heads & 8 \\
    Hidden size & 4,096 \\
    Feed-Forward hidden size & 12,288 \\
    Activation & SILU \\
    RMS norm epsilon & 1e-5 \\
    RoPE $\theta$ & 500k / 20M & ~\usedapache \\
    \bottomrule
\end{tabular}
\end{table}

\subsection{Tokenization}

The Lucie-7B tokenizer decomposes text into subword tokens.
It was trained using Byte-Pair Encoding (BPE) as described by~\cite{bpe}.
We used unicode-level BPE with byte fallback, to avoid out-of-vocabulary,
following the tokenizers of Llama 2, Mistral-7B and CroissantLLM.
The tokenizer was optimized on a subset of the Lucie Training Dataset that includes:
\begin{itemize}
    \item English:
    \begin{itemize}
        \item 1,074 M words from Wikipedia (\url{https://en.wikipedia.org/})
        \item 28 M words from Europarl (English subset)
    \end{itemize}

    \item French:
    \begin{itemize}
        \item 863 M words from Wikipedia (\url{https://fr.wikipedia.org/})
        \item 148 M words from Persée
        \item 94 M words from GallicaMonographies
        \item 25 M words from Europarl (French subset)
    \end{itemize}

    \item Spanish:
    \begin{itemize}
        \item 103 M words from Wikipedia (\url{https://es.wikipedia.org/})
        \item 26 M words from Europarl (Spanish subset)
    \end{itemize}

    \item German:
    \begin{itemize}
        \item 92 M words from Wikipedia (\url{https://de.wikipedia.org/})
        \item 23 M words from Europarl (German subset)
    \end{itemize}

    \item Italian:
        314 M words from Wikipedia (\url{https://it.wikipedia.org/})

    \item Programming Languages:
        198 M words from TheStack
    
\end{itemize}

The vocabulary size is 65,024 tokens,
which is in the order of magnitude of what is recommended by \cite{scaling_tokenization} for a 7B model.
This number is  below the maximum limit to store token indices as \texttt{unsigned int} (65535),
and also a multiple of 256 to optimize matrix multiplication calculations.\footnote{See \url{https://www.thonking.ai/p/what-shapes-do-matrix-multiplications}.}

We implemented the following pre-processing steps (with reverse operations included in the post-processing part of the tokenizer decoder):
\begin{enumerate}
    \item NFC unicode normalization
    \item Removal of control characters \texttt{\textbackslash r} and \texttt{\textbackslash x00}.
    \item Addition of a space not only at the start of a string,
    but also after a set of characters that usually mark the start of sentence, when they are followed by an alpha-numeric character or space:
    \texttt{\textbackslash n}, \texttt{\textbackslash t},
    \texttt{(}, \texttt{[}, \texttt{\{}, \texttt{/}, \texttt{<}, \texttt{'}, \texttt{’}, \texttt{"}, \texttt{«}, \texttt{“}, \texttt{‘}, \texttt{‚}, \texttt{‹}, \texttt{—}, \texttt{–}, \texttt{―}.
\end{enumerate}
There were also constraints imposed on the tokens:
\begin{itemize}
    \item Digits are always isolated (\texttt{123456} is always tokenized as \texttt{1┃2┃3┃4┃5┃6}).
    \item Punctuation marks are always separated from alphabetic characters.
    \item Strings of consecutive spaces up to a fixed arbitrary value are included (from 1 to 8 whitespace characters, from 1 to 4 for tabulations, and from 1 to 2 for carriage returns).
\end{itemize}

\begin{figure}[ht]
    \centering
    \includegraphics[width=\textwidth]{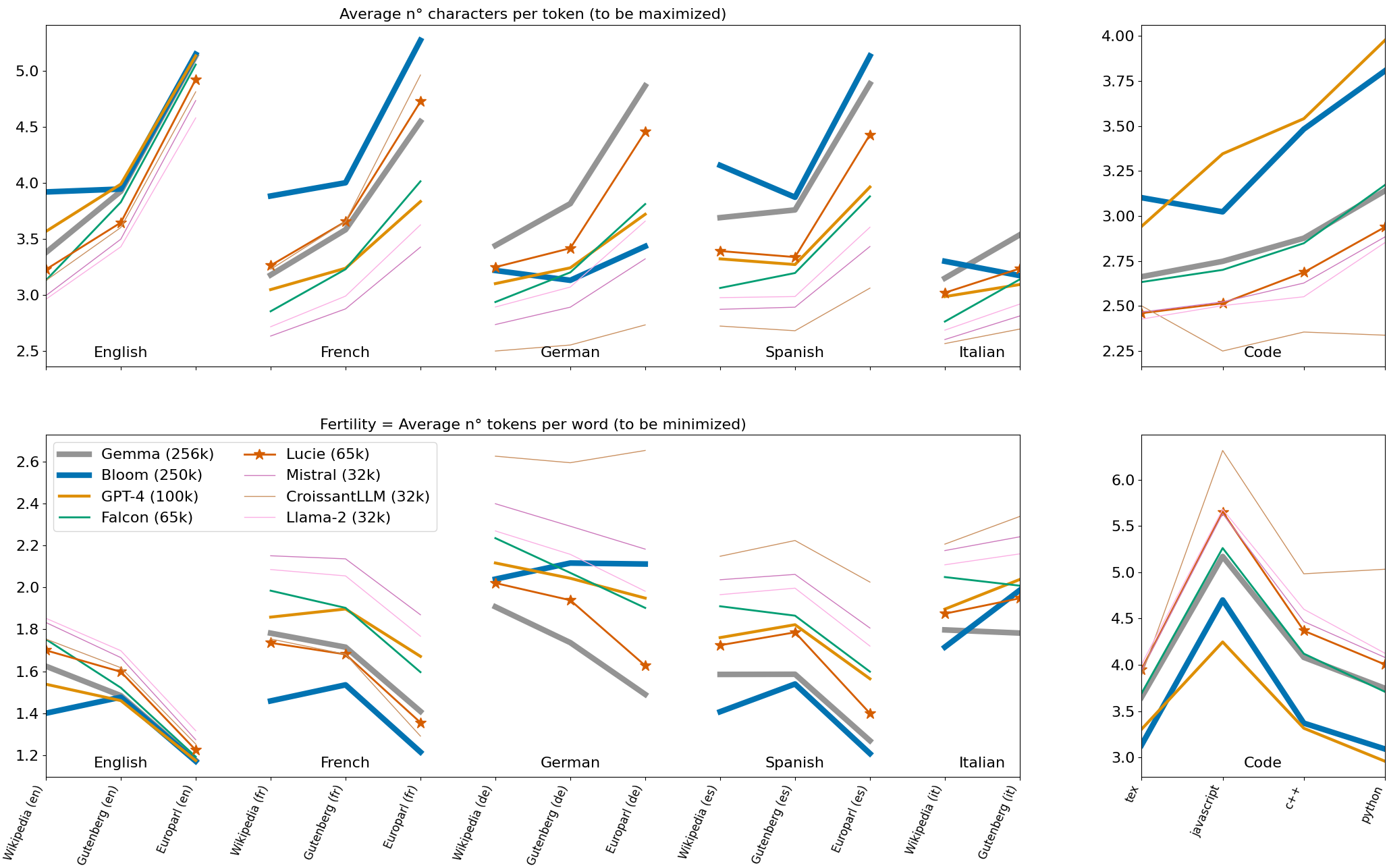}
    \caption{
        Evaluation benchmark results of tokenizers,
        for several languages:
        English, French, German, Spanish and Italian (left), and a few programming languages (right).
    }
    \label{fig:eval_tokenizer}
\end{figure}

In order to evaluate the quality of the tokenizer,
we compared the fertility (average number of tokens per word)
of the Lucie tokenizer with the tokenizers of popular LLMs, on subsets of text samples from:
\begin{itemize}
    \item Wikipedia (a set of samples distinct from those used to train the Lucie tokenizer)
    \item Gutenberg (books)
    \item Europarl (proceedings of the European Parliament)
\end{itemize}
Results are presented in Figure~\ref{fig:eval_tokenizer}
for the different languages modeled by Lucie-7B.
Note that fertility is usually lower on Europarl:
this is due to (1) long words that are represented by a single token
(\textit{e.g.}, ``commission,'' ``européenne,'' ``politique,'' ``monsieur,'' ``président,'' ``parlement,'' and ``également'' in French)
and (2) the fact that there are very few numbers written with digits
(given that most tokenizers decompose numbers into isolated digits, fertility is especially high on numbers).
The empirical results show that the Lucie tokenizer is competitive in terms of fertility with the tokenizers of other state-of-the-art LLMs.
The only tokenizers that exhibit lower fertility  on natural languages are those of Bloom and Gemma,
but these tokenizers have the advantage of larger vocabulary sizes (250k and 256k respectively).
Fertility should be compared for tokenizers that have equal or comparable vocabulary sizes.
On programming languages, tokenizers that can treat many spaces as a unique token have an advantage in terms of fertility (which does not necessarily entail that the LLM built on the tokenizer will offer a better representation of code samples):
this is the case of the GPT-4, Gemma, Bloom and Falcon tokenizers, which 
all show better fertility than the Lucie tokenizer on popular programming languages.
Finally, note that fertility is on the order of 1.5 tokens per word for natural languages, and 4.5 tokens per word for programming languages.

\subsection{Neural network training}
pretraining was carried out in three phases: the main pretraining phase, a phase to extend the context length, and an annealing phase in which the model is trained on high-quality data while the learning rate is decreased linearly to zero. We describe each of these phases below, but first, we begin with an overview of preliminary studies that guided our training choices.

\subsubsection{Preliminary studies}\

Before officially launching the pretraining of Lucie-7B in August 2024, we conducted a number of preliminary studies whose results guided the choices made in the final scripts released with the Lucie models and dataset.
We began in the spring of 2024 by examining state-of-the-art technical reports available at that time, in particular
the Bloom paper~\citep{bloom_short}, website and technical reports~\footnote{\url{https://github.com/bigscience-workshop}}
and the Llama 3.1 technical report~\citep{llama3}. We then conducted a variety of experiments to select the training framework, parallelism configuration and optimal batch size for our training environment on Jean Zay. 

\paragraph{Choosing the right distributed training framework for Jean-Zay.} We implemented and compared preliminary versions of training scripts based on several libraries, in particular FSDP, DeepSpeed, and Megatron-DeepSpeed, adapted to the Jean Zay cluster.
Performance results obtained with FSDP and DeepSpeed, measured relative to select hyper-paremeters 
(context length, number of nodes, etc.),~\footnote{\url{https://gitlab.inria.fr/synalp/plm4all/-/blob/clean/pretraining/deepspeed/eval.md}} indicated that DeepSpeed outperformed FSDP in our case.
Simultaneous experiments with a fork of Megatron-DeepSpeed showed this library to exhibit relatively comparable performances with DeepSpeed.\footnote{\url{https://gitlab.inria.fr/synalp/plm4all/-/tree/clean/pretraining/megatronDS?ref_type=heads}}
Given the fact that Megatron-DeepSpeed had already been used in other available LLM training projects on Jean Zay, in particular Bloom and CroissantLLM, we chose to follow suit and use it for our pretraining. Because Lucie-7B training was carried out on the new H100 partition of Jean Zay, training scripts and slurm parameters released on GitHub are  adapted to the hardware characteristics of the H100 partition of the Jean Zay
cluster as of November 2024.

\paragraph{Choosing the best parallelism configuration.} We employed 3D parallelism, where the model is sharded across tensors (Tensor Parallelism, TP) and layers (Pipeline Parallelism, PP), with the same configuration replicated across the remaining GPUs (Data Parallelism, DP). The value for Data Parallelism is given by \( DP = \frac{TP \times PP}{n_{\text{GPUs}}} \). 

The optimal choice of TP and PP values depends on hardware factors such as inter-node connectivity and other system characteristics. Through experimentation with various parallelism configurations, we observed that certain options fail to achieve optimal training speed. To identify the best configuration, we performed several small-scale training runs on varying numbers of A100 80GB GPUs, testing different parallelism configurations with global batch sizes (GBS) of 192 and 512. For each configuration, we selected the micro-batch size (per GPU batch size) as the largest possible value that avoids out-of-memory errors. 

\begin{table}[h]
\caption{Configurations tested during small-scale training to identify the optimal parallelism strategy. \textit{GBS} and \textit{MBS} stands for Global Batch Size and Micro Batch Size respectively. \textit{TP} and \textit{PP} stand for \textit{Tensor Parallelism} and \textit{Pipeline Parallelism}, respectively.}
\label{tab:p_config}
\centering
\renewcommand{\arraystretch}{1.2}
\setlength{\tabcolsep}{6pt}
\begin{tabular}{@{}ccccccccc@{}}
\toprule
\textbf{nGPUs} & \multicolumn{2}{c}{\textbf{TP=1, PP=1}} & \multicolumn{2}{c}{\textbf{TP=2, PP=1}} & \multicolumn{2}{c}{\textbf{TP=1, PP=2}} & \multicolumn{2}{c}{\textbf{TP=2, PP=2}} \\ 
\cmidrule(lr){2-3} \cmidrule(lr){4-5} \cmidrule(lr){6-7} \cmidrule(lr){8-9}
               & \textbf{GBS} & \textbf{MBS} & \textbf{GBS} & \textbf{MBS} & \textbf{GBS} & \textbf{MBS} & \textbf{GBS} & \textbf{MBS} \\ \midrule
8 GPUs         & 512          & 2            & 512          & 4            & 512          & 2            & 512          & 4            \\ 
               & 192          & 3            & 192          & 4            & 192          & 2            & 192          & 6            \\ \midrule
16 GPUs        & 512          & 2            & 512          & 4            & 512          & 4            & 512          & 4            \\ 
               & 192          & 3            & 192          & 4            & 192          & 4            & 192          & 6            \\ \midrule
32 GPUs        & 512          & 2            & 512          & 4            & 512          & 4            & 512          & 4            \\ 
               & 192          & 3            & 192          & 4            & 192          & 4            & 192          & 6            \\ \midrule
64 GPUs        & 512          & 2            & 512          & 4            & 512          & 4            & 512          & 4            \\ 
               & 192          & 3            & 192          & 4            & 192          & 4            & 192          & 6            \\ \midrule
128 GPUs       & 512          & 2            & 512          & 4            & 512          & 4            & 512          & 4            \\ 
               & 256          & 2            & 192          & 3            & 192          & 4            & 192          & 6            \\ \bottomrule
\end{tabular}
\end{table}

\begin{figure}[h]
    \centering
    \includegraphics[width=0.9\textwidth]{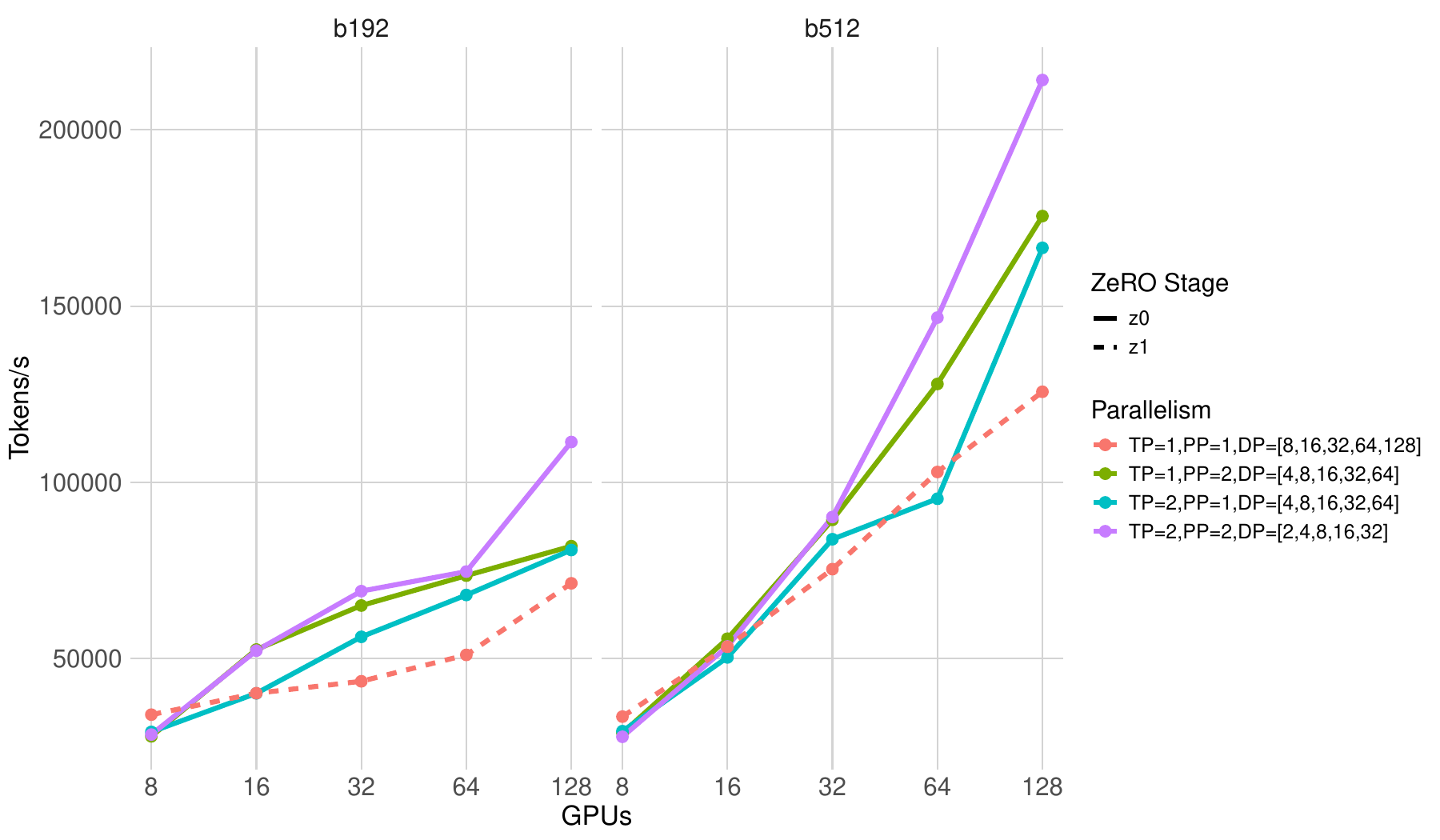}
    \caption{Training speed (tokens per second) as a function of the number of GPUs for different parallelism configurations, with batch sizes of 192 and 512. ZeRO Stage 1 was used to test pure Data Parallelism.}
    \label{fig:parallelism_results}
\end{figure}

To test pure data parallelism, we used ZeRO Optimizer Stage 1, enabling the training model to fit on a single GPU. In ZeRO Optimizer Stage 1, the optimizer state is sharded across GPUs, which frees up GPU memory. Table~\ref{tab:p_config} summarizes the configurations used in these experiments. The results in Figure~\ref{fig:parallelism_results} demonstrate that training speed with pure data parallelism (using ZeRO Optimizer Stage 1) is significantly suboptimal. Conversely, sharding the model across all GPUs without sharding the optimizer state improves training speed. We observed that the best training performance is achieved with a balanced configuration with equivalent TP and PP sizes. In Figure ~\ref{fig:parallelism_results}, we can see that setting TP and PP to 2 increases training speed. Based on these observations, we used TP and PP of size 4 for pretraining on H100 GPUs.

\begin{figure}
    \centering
    \includegraphics[width=0.8\textwidth]{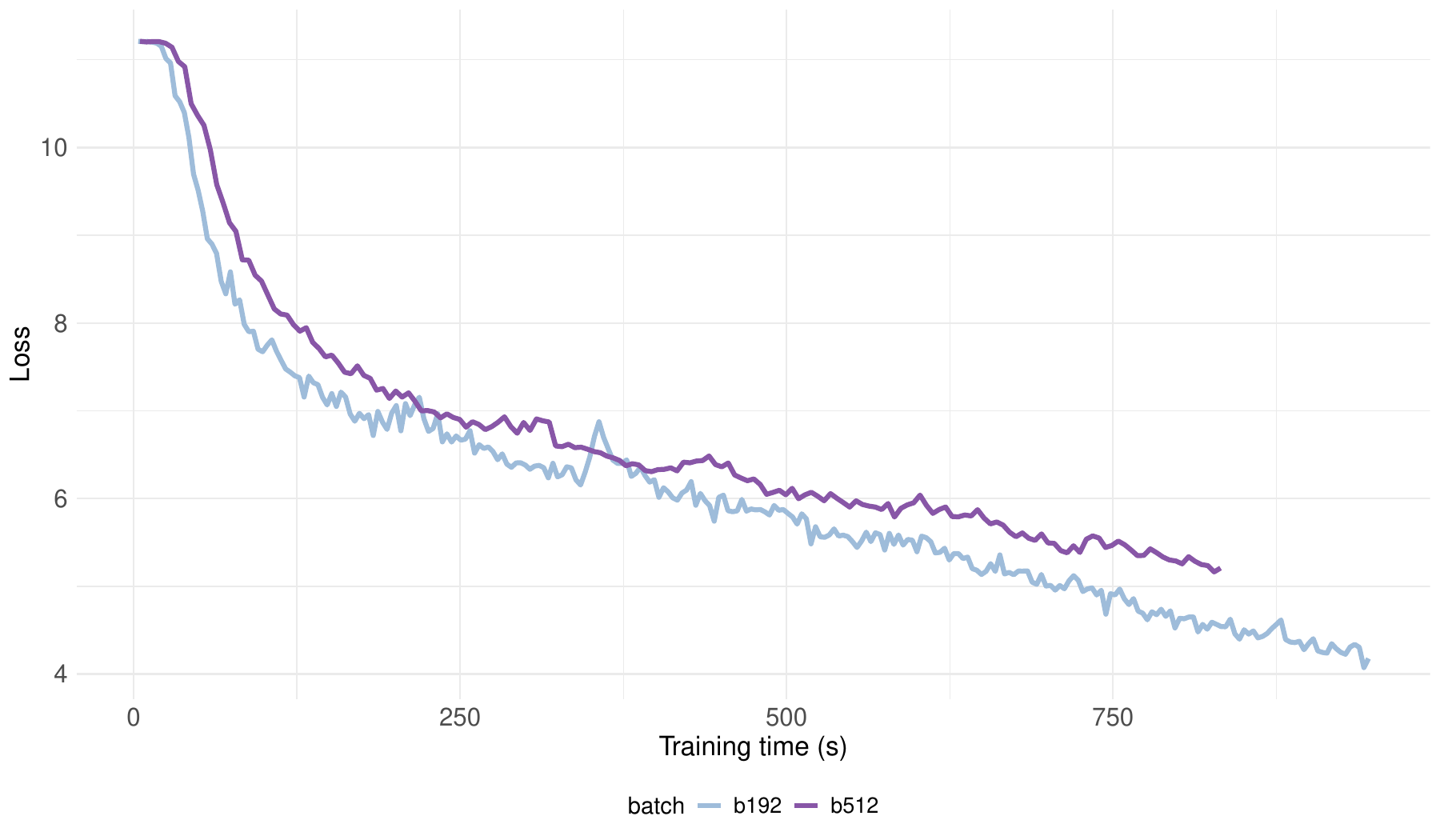}
    \caption{The convergence of training loss as a function of batch size, using TP=2 and PP=2 without optimizer sharding. This setup matches the parallelism configuration that achieved the highest training speed in Figure~\ref{fig:parallelism_results}. }
    \label{fig:time-convergence}
\end{figure}

\paragraph{\textit{Ramp up batch size} as a trade-off between high throughput and convergence speed.} From Figure ~\ref{fig:parallelism_results}, we can see that a larger batch size brings higher training speed. However, we also observed that larger batch sizes slow down the convergence of training loss. Figure~\ref{fig:time-convergence} shows that with the same data and configuration, a higher batch size (512) results in faster training speed but poorer convergence, whereas a batch size of 192 converges better but slows down training speed. As a trade-off, we use  Rampup Batch Size, where the batch size starts at 256 and linearly increases to 1024, with increments of 64 over 10M samples.

\subsubsection{Main pretraining phase}
\label{sec:pretraining1}
The main pretraining phase used 512 H100 80GB GPUs for about 500\,000 GPU hours on the Jean Zay supercomputer.
The initial composition of the training data is shown in Figure~\ref{fig:pie_dataset_composition}.
As explained in Section \ref{sec:data},  some of the datasets were upsampled to balance the training data distribution
yielding the composition for training shown in Figure~\ref{fig:pie_dataset_composition_training},
used in this first phase of Lucie-7B pretraining.

Training hyperparameters, used in torch/Megatron-DeepSpeed, are listed in Table~\ref{tab:training_hyperparameters}.

\begin{table}[H] 
\centering 
\caption{
    Hyperparameters of Lucie-7B training optimization,
    during the first main pretraining phase.
}
\label{tab:training_hyperparameters}    
\begin{tabular}{lr}
    \toprule
    \textbf{Hyperparameter} & \textbf{Value} \\
    \midrule
    Total \# samples (sequences of length 4\,096) & 762M \\
    Total \# tokens & 3.1T \\
    Total \# steps & 753\,851 \\
    \midrule
    Context length & 4\,096 \\
    RoPE $\theta$ & 500k \\
    \midrule
    Initial Batch size & 256 \\
    Final Batch size & 1\,024 \\
    Batch size rampup & by steps of 64 over 10M samples \\
    \midrule
    Learning rate schedule & warmup (2M samples) + cosine annealing \\
    Maximum Learning rate & 3e-4 \\
    Final Learning rate & 3e-5 \\
    \midrule
    Weight decay & 0.1 \\
    Dropout & 0 \\
    Gradient clipping & 1 \\
    Initializer range & 0.009 \\
    \midrule
    Optimizer & AdamW ($\beta_1$=0.9, $\beta_2$=0.95, $\epsilon$=1e-5) \\
    Precision & bfloat16 \\
    Tensor Parallelism (with 512 GPUs) & 4 \\
    Pipeline Parallelism (with 512 GPUs) & 4 \\
    Data Parallelism (with 512 GPUs) & 32 \\
    \bottomrule
\end{tabular}
\end{table}

\subsubsection{Context length extension}
\label{sec:pretraining2}

In order to better accommodate use cases involving tasks like summarization or retrieval augmented generation (RAG), we decided to extend the context length of Lucie-7B to 32,000 tokens. \cite{fu2024dataengineeringscalinglanguage} argue that LLMs are already predisposed to handle context sizes beyond the size of their training sequences and that this capacity can be confirmed through lightweight continual pretraining following the main pretraining phase. The important thing is that the overall domain proportions present in the pretraining data be respected during the context length extension phase. That is, while one might think that it is a good idea to focus on long documents during the extension phase, this can lead to an overrepresentation of certain kinds of documents, like books, that then skew the LLM's probability distribution in undersirable ways. Their approach is to keep domain-level proportions constant but to upsample long documents \textit{within a given domain}.

In line with this approach, we divided the documents of each of our datasets into two categories: documents containing 4096 tokens or fewer and documents containing more than 4096 tokens. We then upsampled the longer documents within each category by a factor of 10 and continued training for 5 billion tokens. During this phase, we decided to further clean certain datasets that were used in pretraining. This led to the version 1.2 of the Lucie Training Dataset described in Section \ref{sec:data-preparation}.

Training hyperparameters are the same as during the first pretraining phase (Table~\ref{tab:training_hyperparameters}),
except for the ones mentioned in Table~\ref{tab:training_hyperparameters_context_extension}.

\begin{table}[ht] 
\caption{
    Hyperparameters of Lucie-7B training optimization,
    during the second phase of context length extension.
}
\label{tab:training_hyperparameters_context_extension}
\centering 
\begin{tabular}{lr}
    \toprule
    \textbf{Hyperparameter} & \textbf{Value} \\
    \midrule
    Total \# samples & 156\,250 \\
    Total \# tokens & 5B \\
    Total \# steps & 1\,220 \\
    \midrule
    Context length & 32\,000 \\
    RoPE $\theta$ & 20M \\
    \midrule
    Batch size & 128 \\
    Learning rate & 2e-5 \\
    Learning rate schedule & constant \\
    \midrule
    Tensor Parallelism (with 128 GPUs) & 4 \\
    Pipeline Parallelism (with 128 GPUs) & 4 \\
    Data Parallelism (with 128 GPUs) & 8 \\
    \bottomrule
\end{tabular}
\end{table}

\subsubsection{Annealing with high-quality data}
\label{sec:pretraining3}

We finished the pretraining with an annealing phase, focusing on high-quality datasets in order to boost the model's performance.
The hyperparameters used during this phase were the same as during the context extension phase (see Table~\ref{tab:training_hyperparameters_context_extension}), except for the learning rate schedule, which was linearly annealed from $3e-5$ to $0$.

During the annealing phase, we adjusted the data mix presented in Figure~\ref{fig:pie_dataset_composition_training}, and put more emphasis on high-quality and math-related datasets.
We monitored the performance with evaluation on the GSM8K (Grade School Math 8K) benchmark, following \cite{llama3}.

We introduced four new datasets during this phase:
\begin{itemize}
    \item OpenWebMath. \footnote{\url{https://huggingface.co/datasets/open-web-math/open-web-math}} A filtered subset of Common Crawl focusing on math-related datasets. This replaced the general web datasets from the earlier training phases.
    \item StackMathQA. \footnote{\url{https://huggingface.co/datasets/math-ai/StackMathQA}} A curated dataset extracted from Stack Exchange sites. It includes 2 million question-answer pairs. 
    \item Flan v2. \footnote{\url{https://huggingface.co/datasets/Open-Orca/FLAN}} We followed OLMO 2 preprocessing \citep{olmo} but expanded the dataset size, using 50 times more data samples.
    \item Aya Collection. \footnote{\url{https://huggingface.co/datasets/CohereForAI/aya_collection}} A multilingual dataset containing human-made instructions in multiple languages, translated instruction datasets and new templated datasets in multiple languages~\citep{singh2024aya}.
\end{itemize}

To identify the optimal data mix, we performed a brief ablation study testing two main parameters:
\begin{itemize}
\item Number of training tokens. We compared two configurations with 40M and 5B training tokens.
\item Data mix proportions. We tested four data mixes for the 40M training token configuration and six data mixes for the 5B configuration.
\end{itemize}
Our results indicated that the best strategy was to use more math-related datasets and to introduce the Flan v2 dataset. The final data mix proportions are detailed in Table~\ref{tab:annealing_mix}.

\begin{figure}[H]
    \centering
    \includegraphics[width=10cm]{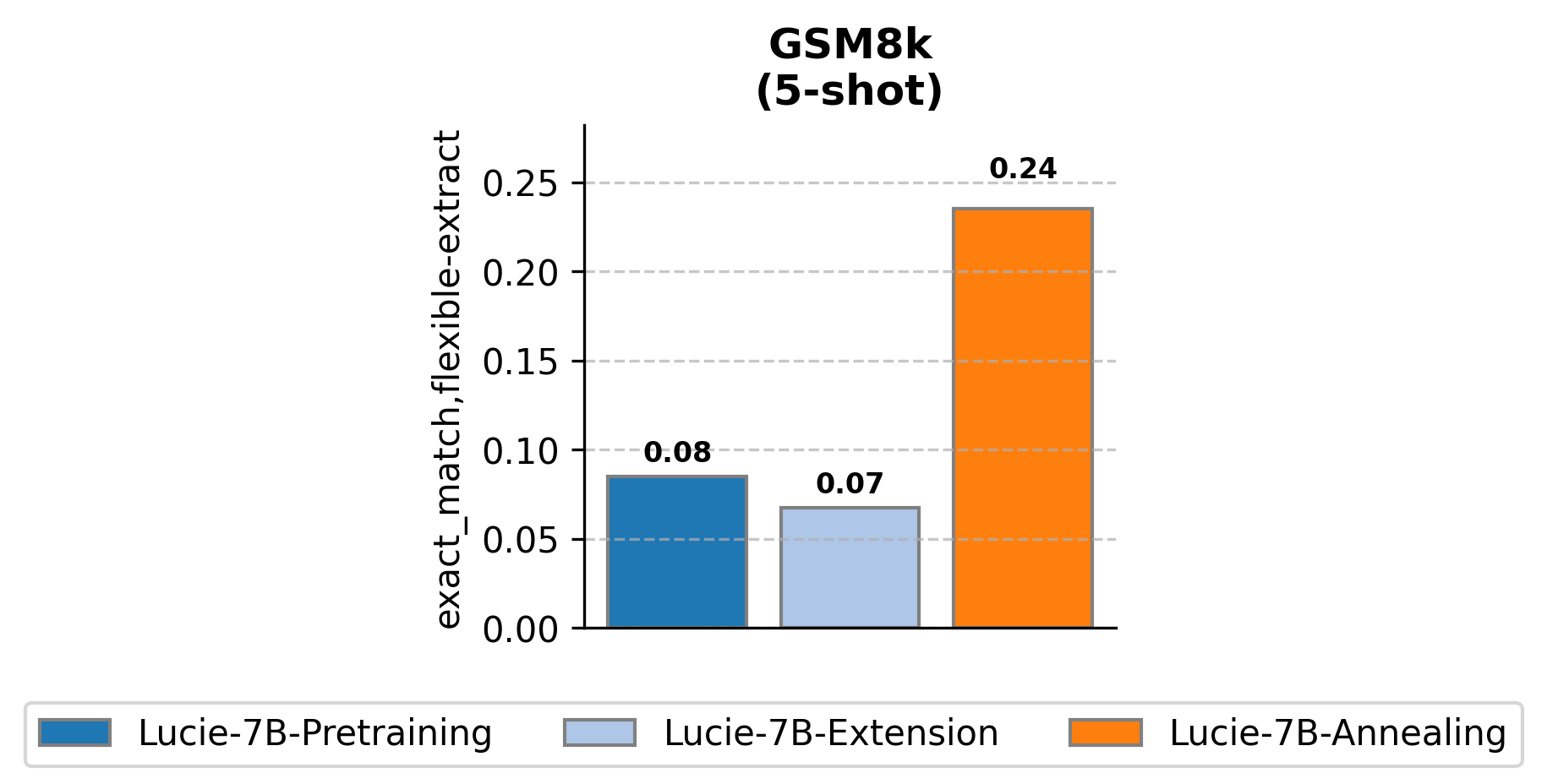}
    \caption{
        Evolution of GSM8K metric for each stage.
    }
    \label{fig:annealing_eval}
\end{figure}

Figure~\ref{fig:annealing_eval} illustrates the GSM8K metric performance for the three pretraining phases. The accuracy improves significantly from 8.4\% after the initial phase (6.7\% after the context extension phase) to 23.5\% after the annealing phase. However, these scores remain lower than state-of-the-art models suggesting a lack of math-related data in our training dataset.

\begin{table}[H]
    \caption{
        Final data mix used during the annealing phase. 
    }
    \label{tab:annealing_mix}
    \centering 
    \begin{tabular}{lll}
        \toprule
        \textbf{Dataset} & \textbf{Languages} & \textbf{Proportion} \\
        \midrule
        OpenWebMath  & en & 0.2 \\
        Pes2o & en & 0.1 \\
        MathPile & en & 0.25 \\ 
        StackMathQA   & en & 0.05 \\
        Flan v2  & en & 0.2 \\
        Wikipedia & en, fr, es, it, de & 0.1 \\
        Aya Collection   & fr, es, it, de & 0.1 \\
        \bottomrule
    \end{tabular}
\end{table}

\subsection{Environmental impact}
When assessing the environmental impact of an AI model, it is essential to consider both the training phase and subsequent inference costs. While the training phase is the most resource-intensive, the inference phase can have a greater overall impact once applications built on the model achieve widespread adoption. We thus provide information on both phases here.

\paragraph{Training Phase.}
As the Jean Zay supercomputer is located in France, it runs on low-carbon nuclear electricity. Our primary training took place between September and December 2024, during a period of low water stress, when the French energy mix was not reliant on Germany, where electricity production is not carbon-free.
CNRS/IDRIS has implemented a waste heat recovery system that provides heating for all buildings on the Saclay Plateau. The Power Usage Effectiveness (PUE) for the H100 partition is 1.21.
According to GENCI, the carbon footprint coefficient for this H100 partition is 25.7 g CO$_2$-eq per GPU-hour. Given that our training consumed 500,000 GPU hours, this results in a total carbon footprint of 12.85 metric tons of CO$_2$ equivalent—compared to 31.22 metric tons CO$_2$-eq for LLaMA 2 7B or 390 metric tons CO$_2$-eq for LLaMA 3 8B.

\paragraph{Inference Phase.}
We currently do not have quantified data on the electricity consumption of an inference infrastructure based on Lucie 7B, using vLLM, for example. However, several key points should be noted:
\begin{itemize}
    \item Smaller models are less energy-intensive as they require fewer GPU and memory resources.
    \item Quantized versions of the instruct models are available and can run on most desktop PCs without a GPU using tools like Ollama.
\end{itemize}
For our initial commercial deployments, we have partnered with Exaion, a company that offers an infrastructure based on recycled A4000 GPUs.

\subsection{Evaluation}

\subsubsection{Convergence and learning curves}

\begin{figure}[H]
    \centering
    \includegraphics[width=\linewidth]{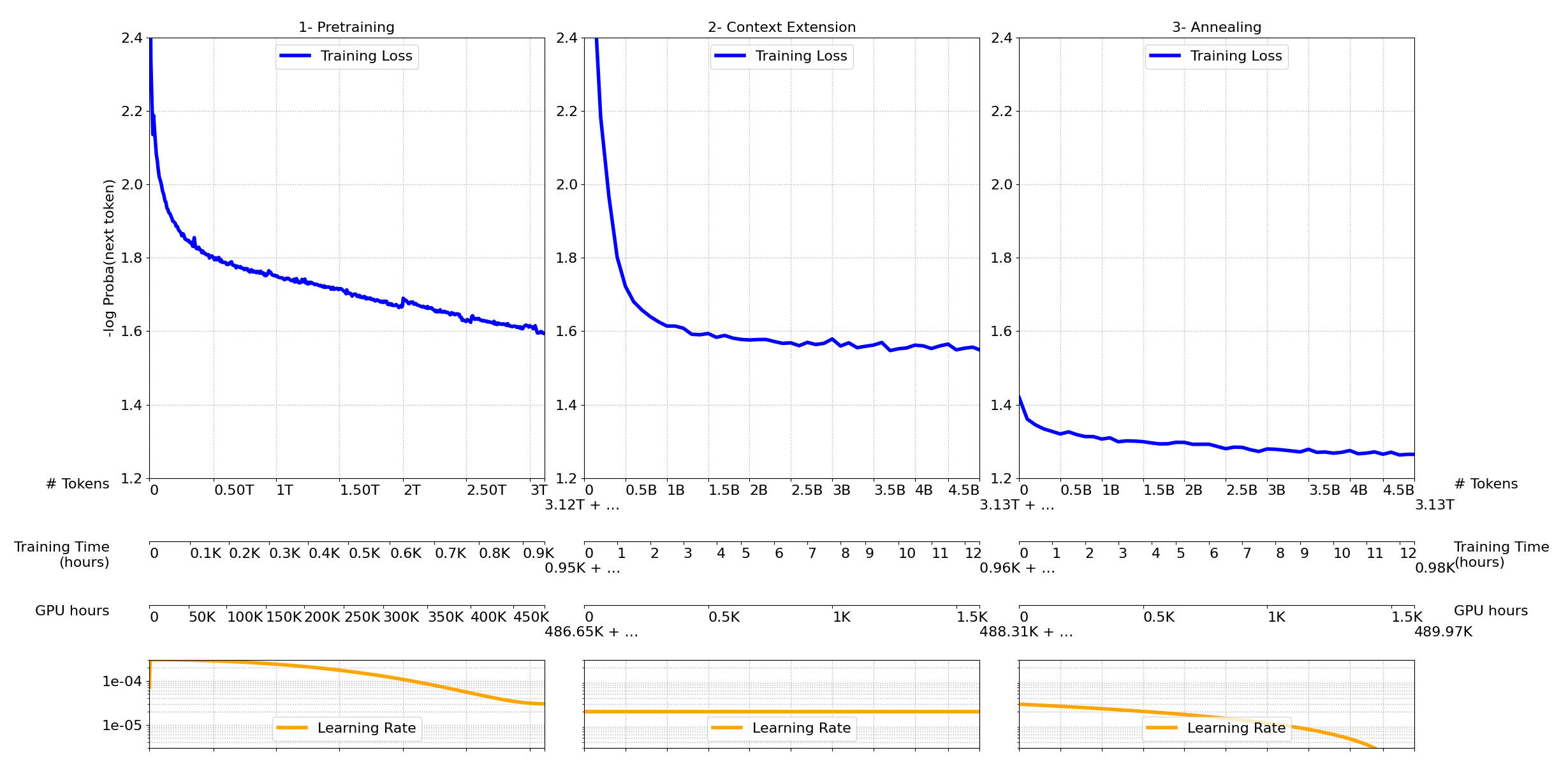}
    \caption{
        Convergence curve of Lucie-7B: evolution of training loss during the three pretraining phases.
    }
    \label{fig:convergence-curve}
\end{figure}

Figure~\ref{fig:convergence-curve} shows the convergence curve of Lucie-7B;
that is, the evolution of the training loss during the three pretraining phases.
We did not observe any instability during training, nor did we need to intervene at any point. The training loss curve during the main training phase, along with the trends in Figures~\ref{fig:learning-curve-evaluation-english},\ref{fig:learning-curve-evaluation-french},\ref{fig:learning-curve-evaluation-multilingual}, suggests that the model continued to benefit from training even after processing 3T tokens. This indicates that further training could have led to additional performance improvements.

\subsubsection{Benchmark evaluations}\label{sec:bench-pretraining}
We evaluate Lucie-7B on a variety of well-known LLM benchmarks designed to target different types of LLM skills. We include not only results for the final Lucie-7B model after annealing but also results for intermediate checkpoints, measured every 25,000 steps (approximately every 100 billion tokens) during the pretraining phase. 

\begin{figure}[h]
\centering
\includegraphics[width=\linewidth]{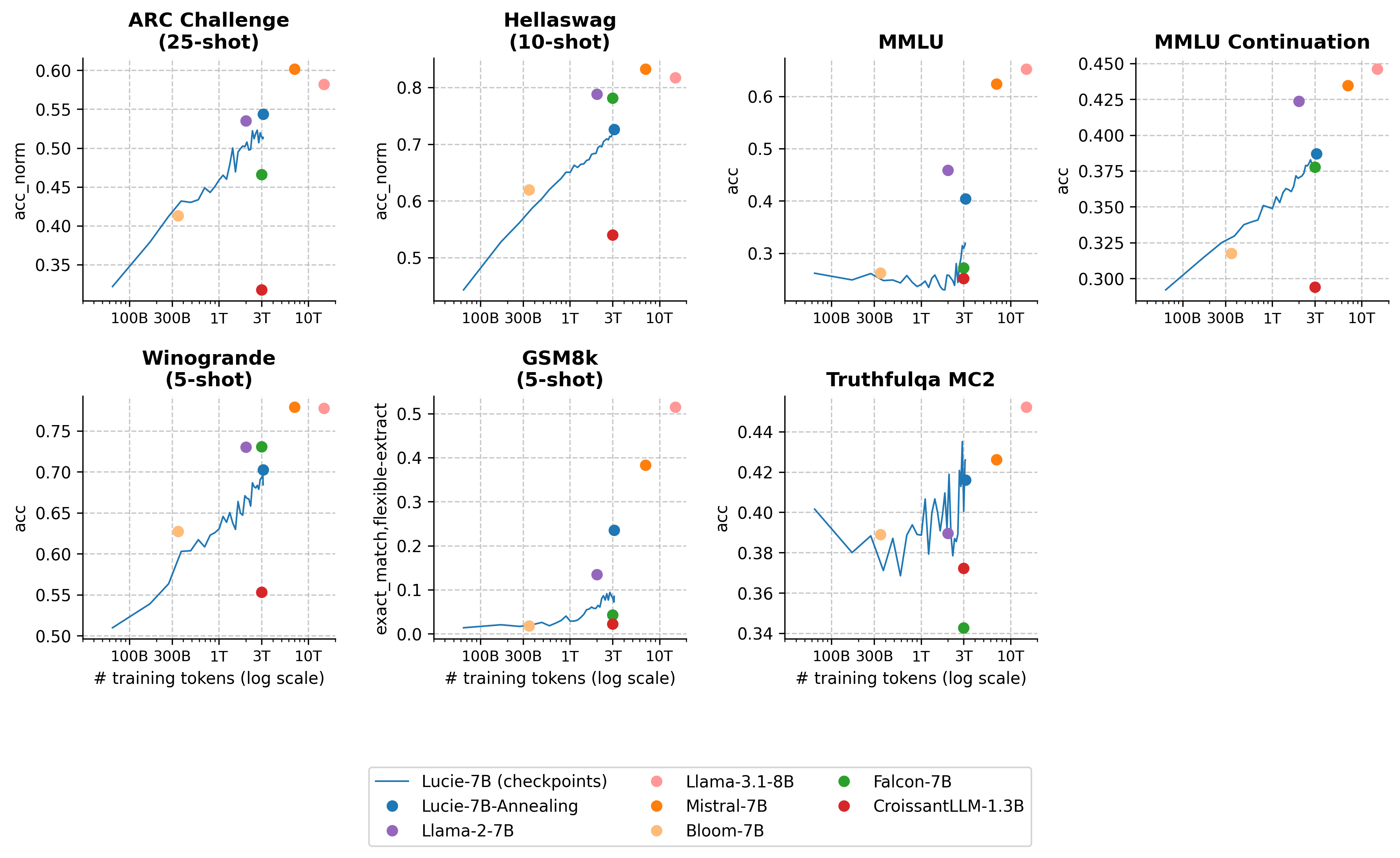}
\caption{
    Evaluation of baseline LLM and Lucie-7B checkpoints on English NLP benchmarks.
}
\label{fig:learning-curve-evaluation-english}
\end{figure}

\begin{figure}[h]
\centering
\includegraphics[width=\linewidth]{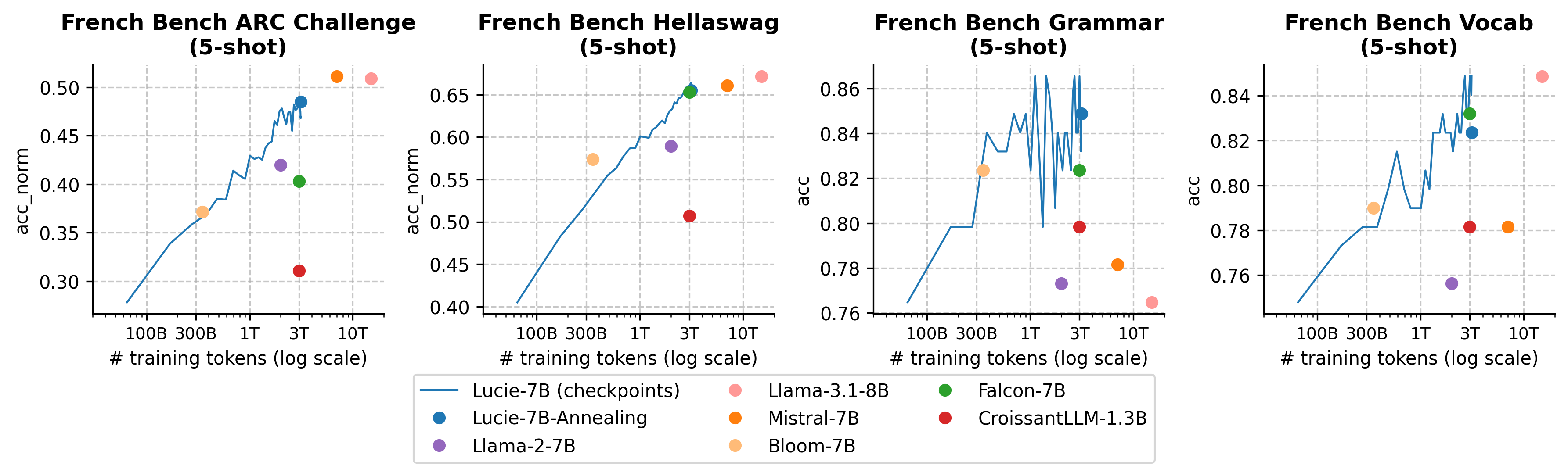}
\caption{
    Evaluation of baseline LLM and Lucie-7B checkpoints on French NLP benchmarks.
}
\label{fig:learning-curve-evaluation-french}
\end{figure}

\begin{figure}[h]
\includegraphics[width=\linewidth]{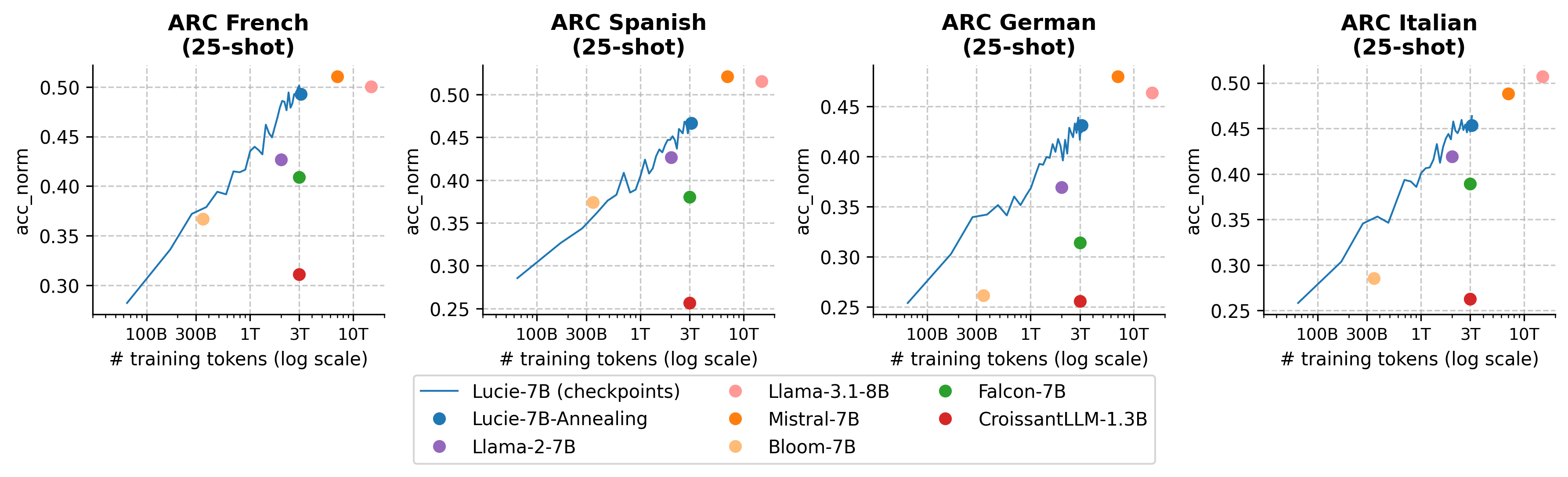}
\caption{
    Evaluation of baseline LLM and Lucie-7B checkpoints on NLP benchmarks in several European languages.
}
\label{fig:learning-curve-evaluation-multilingual}
\end{figure}

\paragraph{Selected benchmarks.}
We use Eleuther AI's lm-eval-harness library for evaluation~\citep{eval-harness}\footnote{\url{https://github.com/EleutherAI/lm-evaluation-harness/tree/main}} and select three main categories of benchmarks to assess Lucie-7B's capacities.
\begin{itemize}
    \item \textbf{The Hugging Face Open LLM Leaderboard}. \footnote{\url{https://huggingface.co/spaces/open-llm-leaderboard-old/open_llm_leaderboard}} An English benchmark for diverse tasks that includes: ARC Challenge \citep{clark2018thinksolvedquestionanswering}, HellaSwag \citep{zellers2019hellaswagmachinereallyfinish}, MMLU \citep{hendrycks2021measuringmassivemultitasklanguage}, WinoGrande \citep{sakaguchi2019winograndeadversarialwinogradschema}, GSM8k \citep{lin2022truthfulqameasuringmodelsmimic}, and TruthFulQA \citep{lin2022truthfulqameasuringmodelsmimic}. Notably, for MMLU, we evaluate using two formats: the classic approach of predicting the index letter and the ``continuation'' approach, where we use the full answer as the target. The continuation format is less abstract and benefits early-stage checkpoints. Results for these tasks, including both configurations of MMLU, are shown in Figure \ref{fig:learning-curve-evaluation-english}.
    \item \textbf{FrenchBench}. A French benchmark introduced in ~\cite{croissant} comprising multiple tasks designed to evaluate different aspects of performance in French. In this section, we focus on four tasks from the FrenchBench Multiple Choice suite: French Language Test (covering grammar and vocabulary), French Hellaswag, and French ARC-Challenge. For each of these tasks, we use 5 few-shot examples to assess model performance. Results for these tasks are shown in Figure \ref{fig:learning-curve-evaluation-french}.
    \item \textbf{Multilingual ARC Challenge}. Introduced in \cite{lai2023okapiinstructiontunedlargelanguage}, this benchmark includes translated versions of the ARC Challenge in several languages. We focus on evaluating model performance in four non-English languages from the Lucie-7B pretraining corpus: French, Spanish, German, and Italian. For consistency with the English ARC Challenge leaderboard, we use 25 few-shot examples for these evaluations. Results for these tasks are shown in Figure \ref{fig:learning-curve-evaluation-multilingual}.
\end{itemize}

\paragraph{Comparison models.}
We compare Lucie-7B checkpoints and the final Lucie-7B pretrained model after annealing with five other open-weight foundation models of similar parameter size (around 7 billion parameters). We also add CroissantLLM, a much smaller model, due to the fact that its datamix is the closest to that of Lucie-7B and it was trained on a similar quanity of tokens.\footnote{We note that token comparisons are only approximate due to the fact that each model uses its own tokenizer.} 
\begin{itemize}
    \item \textbf{Llama-2-7B} \citep{llama2}. This is the second iteration of the Llama foundation models and was trained on roughly 2 trillion tokens. We include it in our comparisons because its architecture and size are very similar to that of the more recent Llama-3.1-8B but it was trained on far fewer tokens, offering an idea of how dataset size can impact performance.
    \item \textbf{Llama-3.1-8B} \citep{llama3}. This is the latest iteration of the Llama series from Meta, trained on 15 trillion tokens. It contains slightly more parameters than that of Llama-2-7B, mainly due to a larger vocabulary size of 128,000 tokens. We include it due to its state of the art performance.
    \item \textbf{Mistral-7B} \citep{jiang2023mistral}. We are unaware of the size or composition of the Mistral-7B training set. We include it as an example of a strong model also coming out of France.
    \item \textbf{Bloom-7B} \citep{bloom_short}. An early major contribution to efforts to create fully open LLMs, Bloom-7B is a massively multilingual model. As it was trained on only 300 billion tokens, comparison with this model also provides an idea of the impact of dataset size. 
    \item \textbf{Falcon-7B} \citep{almazrouei2023falconseriesopenlanguage}. Falcon-7B is trained on  four of the five languages on which Lucie-7B was trained, although on only about half the amount of tokens.
    \item \textbf{CroissantLLM} \citep{croissant}. Like Lucie-7B, CroissantLLM was trained on equal quantities of French and English data and on around 3 trillion tokens.  Note that CroissantLLM is a significantly smaller model with only 1.3 billion parameters, allowing us to analyze the scalability of Lucie-7B.
\end{itemize}

\paragraph{Results.}

Figures~\ref{fig:learning-curve-evaluation-english},~\ref{fig:learning-curve-evaluation-french}~and~\ref{fig:learning-curve-evaluation-multilingual}
show how the performance of Lucie-7B evolved on standard benchmarks during pretraining. 
Results exhibit a logarithmic trend and do not reach a plateau, suggesting that the model could benefit from extended training.

Compared to CroissantLLM and Bloom, Lucie-7B is consistently superior across all benchmarks. This underscores the scalability of the Lucie-7B model along two key dimensions: model size, when compared to CroissantLLM, and training token count, when compared to Bloom.

While Falcon-7B has a lower token count that Lucie-7B, however, here the comparison is more mixed. Lucie-7B has better performance on ARC Challenge for all languages but not on Hellaswag in English or French or on Winogrande in English.

For multilingual benchmarks, the evolution of Lucie-7B's performance places it on a trajectory with the strongest models, Llama-3.1-8B and Mistral-7B, with Lucie-7B even outperforming these models on certain benchmarks. 

On English benchmarks, though, Llama-3.1-8B and Mistral-7B clearly pull ahead. A factor that might have negatively impacted Lucie-7B's performance on English benchmarks is the fact that its training dataset had an equal balance of French and English data, but this hypothesis would need to be further explored.

Lucie-7B's mathematical reasoning remains a weak point. On GSM8K, a clear performance gap emerges between models trained on fewer than 3T tokens and those trained on larger datasets (with the caveat that we are not aware of the size of  Mistral-7B's training dataset).  As discussed in Section~\ref{sec:pretraining3}, however, performance on this benchmark was improved during the annealing phase by adjusting the data mix to emphasize mathematical content.

Overall, Lucie is currently more comparable to Llama 2 than Llama 3. This underscores the fact that model performance is not only linked to the quantity of pretraining data used, as Llama 2 is trained on fewer tokens than Lucie, but also a matter of the nature of the data.  A deeper investigation into data mix composition might help improve overall performance. In future work, we plan to study how training on equal amounts of French and English data might have impacted Lucie's results on English benchmarks. We will also aim to better understand how the type and quality of our French data---our French datasets contain more documents retrieved from OCR for example and less refined web data---might have influenced performance.  

Finally, a notable limitation on the benchmarks discussed in this section is that many of them were created by translating English datasets, making it challenging to evaluate cultural biases. Addressing this issue will be a key focus of future work within the OpenLLM project.

\subsubsection{Needle in a haystack}\label{sec:needle-pretraining}

To test the efficacy of the context length extension phase, we appealed to the ``needle in a haystack'' benchmark~\citep{needle_haystack}, which is designed to evaluate the ability of large language models (LLMs) to retrieve information from a long context.

The benchmark involves placing a short sentence (the ``needle'') within a long distractor document (the ``haystack''). The model is then evaluated on its ability to retrieve the needle within the haystack. The needle is placed at varying depths within the haystack, and the haystack's length is adjusted to represent both short and long contexts.

We used the implementation from \cite{fu2024dataengineeringscalinglanguage},\footnote{\url{https://github.com/FranxYao/Long-Context-Data-Engineering}} where the needle is the sentence: ``The best thing to do in San Francisco is eat a sandwich and sit in Dolores Park on a sunny day.'' The haystack consists of an excerpt from a Paul Graham essay. The model is then prompted with the instruction: ``The best thing to do in San Francisco is'' and is expected to respond with: ``eat a sandwich and sit in Dolores Park on a sunny day.''

\begin{figure}[H]
    \centering
    \begin{minipage}{0.45\textwidth}
        \centering
        \includegraphics[width=\textwidth]{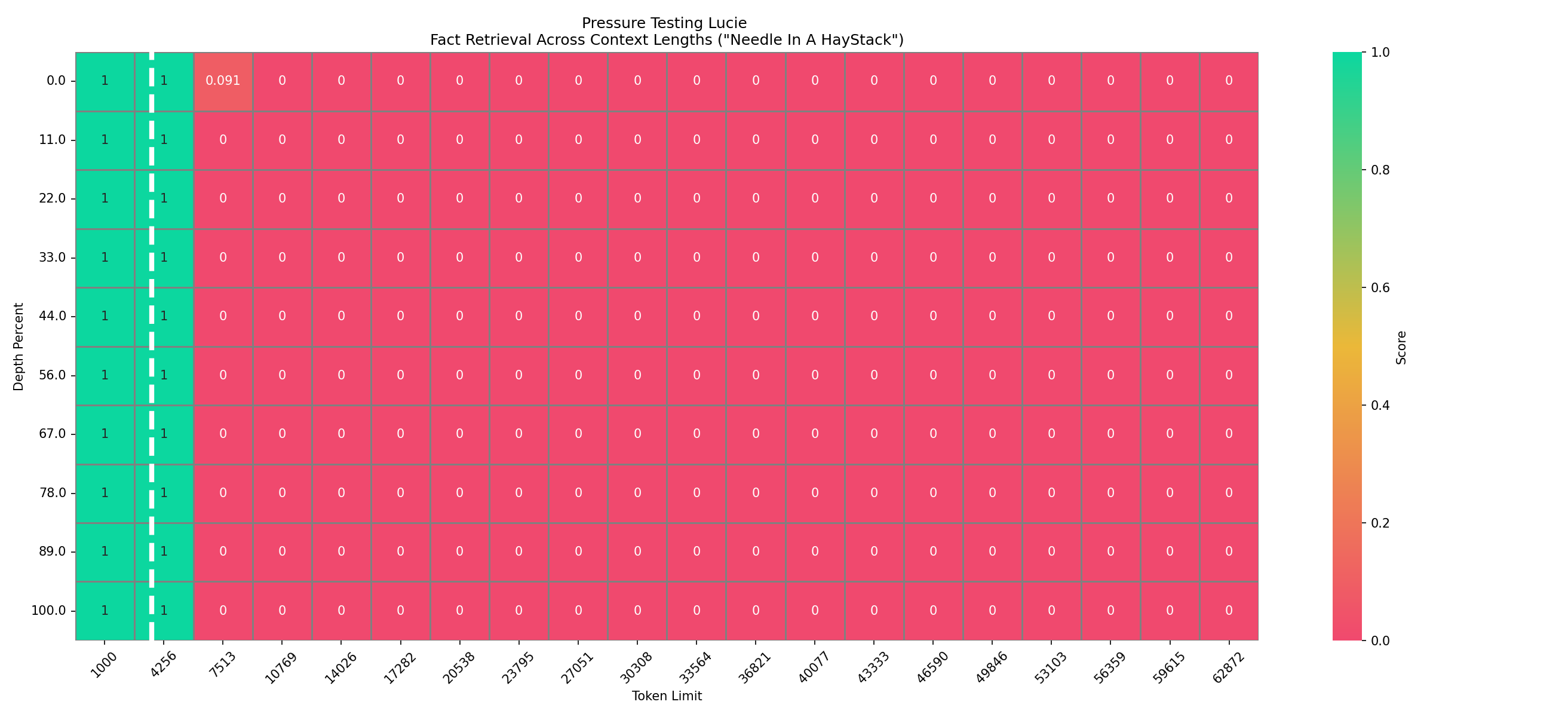}
        \caption{
            Effective context window size for Lucie-7B after the first phase of pretraining,
            measured with the needle in a haystack method \cite{needle_haystack}.
        }
        \label{fig:needle_a}
    \end{minipage}
    \hfill
    \begin{minipage}{0.45\textwidth}
        \centering
        \includegraphics[width=\textwidth]{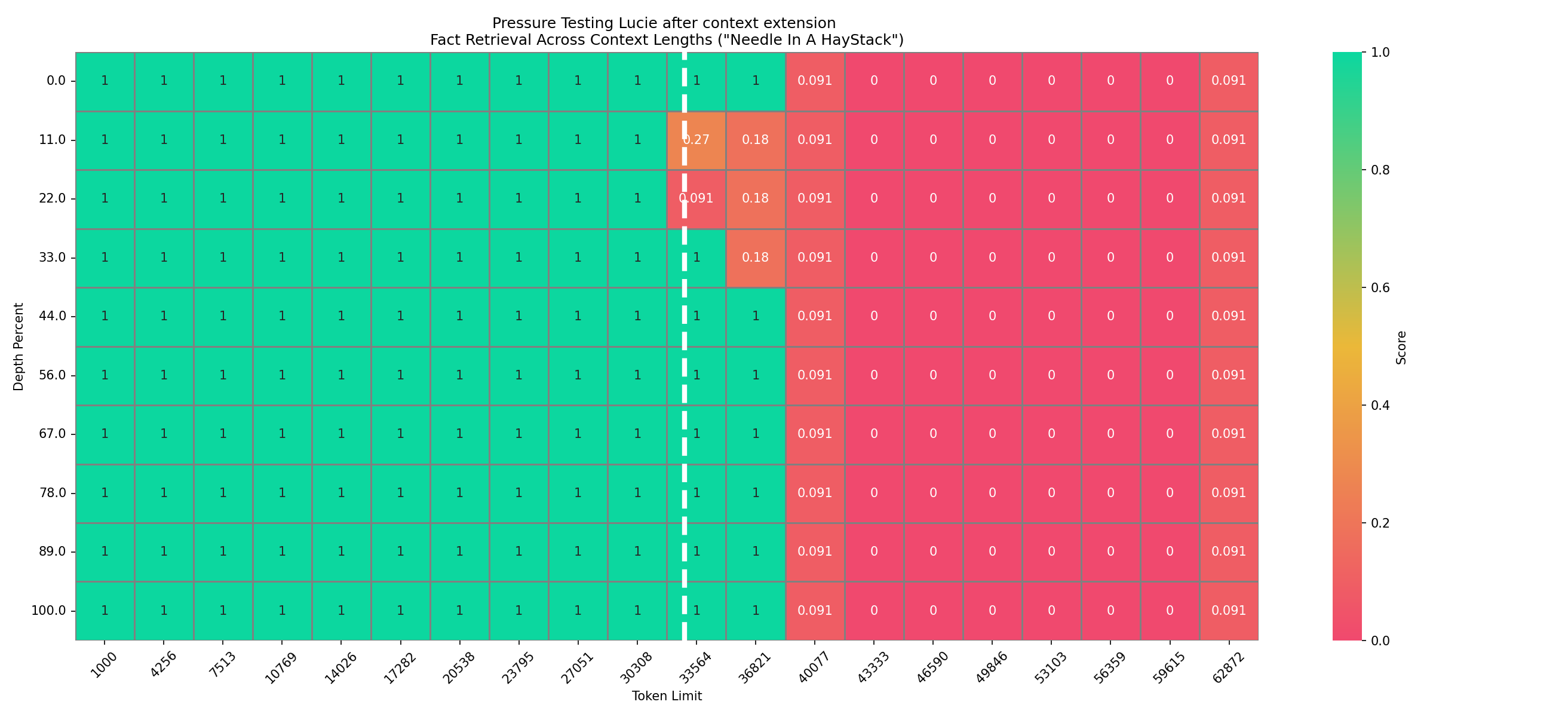}
        \caption{
            Effective context window size for Lucie-7B after context length extension (2nd phase).
        }
        \label{fig:needle_b}
    \end{minipage}
    \\
    \vspace{30mm}
\end{figure}

\begin{figure}[H]
\begin{center}

    \begin{minipage}{0.45\textwidth}
        \centering
        \includegraphics[width=\textwidth]{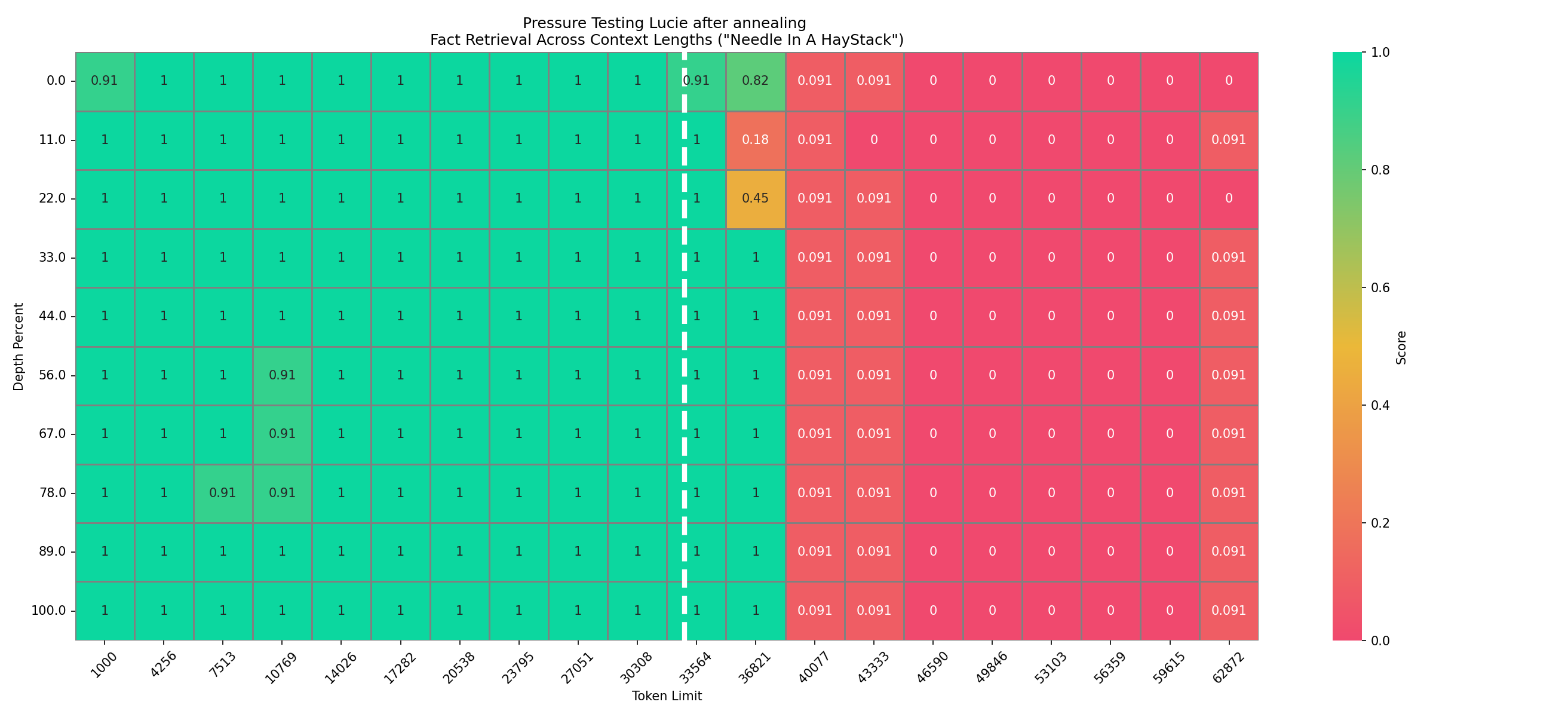}
        \caption{
            Effective context window size for Lucie-7B after annealing (3rd phase).
        }
        \label{fig:needle_c}
    \end{minipage}
    \end{center}
\end{figure}

Results of this benchmark are shown in Figures~\ref{fig:needle_a},~\ref{fig:needle_b}~and~\ref{fig:needle_c}. In Figure \ref{fig:needle_a}, the model can retrieve information at all depths (y-axis) up to a context length of around 4,256 tokens as everything to the left of the white line is green. However, the model fails to find the needle in larger contexts, as indicated by the large red region.

After context extension (Figure~\ref{fig:needle_b}), the target context length is 32,000 tokens. The model finds the needle up to a context length of nearly 34,000 tokens at almost all depths with two exceptions (between the top 10-30\% of the haystack for a context length of over 32,000) as indicated by the two red-orange squares to the left of the white dashed line. The results are even better after annealing, where the model finds the needle at all depths up to 34,000 tokens.

\subsection{Pretrained foundation model release}

Lucie-7B is available on Hugging Face at \url{https://huggingface.co/OpenLLM-France/Lucie-7B}.
This repository also makes available, in revision versions, intermediate checkpoints, after regular training steps\footnote{
    Checkpoints from the first main pretraining phase are available for training steps:
    5\,000,
    10\,000,
    \ldots,
    25\,000,
    50\,000,
    75\,000,
    100\,000,
    \ldots,
    750\,000,
    753\,851.
    Checkpoints from the second main pretraining phase are available for training steps:
    250, 500, 750, 1000, 1220.
}.
While those checkpoints are in \texttt{transformers}\footnote{\url{https://huggingface.co/docs/transformers}} format,
full checkpoints with optimizer states are available
in the \texttt{Megatron-DeepSpeed}\footnote{\url{https://github.com/microsoft/Megatron-DeepSpeed}}/\texttt{pytorch}\footnote{\url{https://pytorch.org/}} format
at \url{https://huggingface.co/OpenLLM-France/Lucie-7B-optimizer-states} (universal checkpoints)
and \url{https://huggingface.co/OpenLLM-France/Lucie-7B-optimizer-states-512GPU} (original checkpoints with 512-GPU parallelizing).

The pretrained foundation model Lucie-7B can be loaded and used to infer an answer with the following code.
In this example, 1-shot prompting is used to answer a question in French
(``Quelle est la capitale de la France~?'' --~``What is the capital of France?'').\\

\hspace{5mm} \begin{minipage}[b]{0.8\textwidth}
{\small
\begin{minted}[linenos]{python}
import transformers

model_name = "OpenLLM-France/Lucie-7B"

tokenizer = transformers.AutoTokenizer.from_pretrained(model_name)
model = transformers.AutoModelForCausalLM.from_pretrained(model_name,
    device_map="auto",
    load_in_4bit=True       # For efficient inference, if quantization is supported by the GPU card
)

# Wrap the model in a text generation pipeline
pipeline = transformers.pipeline("text-generation", model=model, tokenizer=tokenizer)

generation_kwargs = dict(
    num_return_sequences=1,               # Number of variants to generate.
    return_full_text= False,              # Do not include the prompt in the generated text.
    do_sample=True,
    temperature=1.0, top_p=1, top_k=None, # Sampling parameters.
    max_new_tokens=200,                   # Maximum length for the output text (in number of tokens).
)

# Try 1-shot question answering

prompt = """\
Quelle est la capitale de l'Espagne ? Madrid\n\
Quelle est la capitale de la France ?\
"""
completions = pipeline(prompt, **generation_kwargs)
for completion in completions:
    print(prompt + " […]" + completion['generated_text'])
\end{minted}
\begin{alltt}
Quelle est la capitale de l'Espagne ? Madrid
Quelle est la capitale de la France ? […] Paris
Quelle est la capitale de l'Italie? Rome
Quelle est la capitale de la Grande-Bretagne? Londres
Quelle est la capitale de la Suisse? Berne
Quelle est la capitale du Portugal? Lisbonne
Quelle est la capitale de l'Algérie? Alger
...    
\end{alltt}
}

\end{minipage}

\section{After pretraining: Instruction fine-tuning}\label{sec:instruction}
A pretrained generative model has learned to predict, when given an input sequence or ``prompt,'' the most likely next word in the sequence. The goal in the pretraining phase is to expose the model to a wide variety of open-ended data so that it learns which words tend to occur with which words. For instance, it should learn that \textit{red} and \textit{wine} are likely to occur together but it should also learn to place lower probability on a co-occurrence of \textit{blue} and \textit{wine}. During the instruction phase, the point is to guide the model to predict certain types of sequences given certain types of prompts: it should provide an answer when asked a question, provide a summary of an input document when requested, provide a list when asked for a recipe, and so on. Instruction tuning is a kind of multitask supervised fine-tuning (SFT) in which the model is trained on pairs of a variety of input instructions with sample responses.  
Loss during training for instruction tuning can optionally be calculated on both the instruction and response, or only on the assistant's response. 

While the principal contributions of our project are the Lucie Training Dataset and the Lucie-7B foundation model, we decided to also release two instruction-tuned models, Lucie-7B-Instruct-human-data\footnote{\url{https://huggingface.co/OpenLLM-France/Lucie-7B-Instruct-human-data}} and Lucie-7B-Instruct-v1.1.\footnote{\url{https://huggingface.co/OpenLLM-France/Lucie-7B-Instruct-v1.1} Note that the original release of Lucie-7B was accompanied by a model called Lucie-7B-Instruct that was fine-tuned on: Alpaca-cleaned \url{https://huggingface.co/datasets/yahma/alpaca-cleaned}, Alpaca-cleaned-fr \url{https://huggingface.co/datasets/cmh/alpaca_data_cleaned_fr_52k}, Magpie-Gemma \url{https://huggingface.co/datasets/Magpie-Align/Magpie-Gemma2-Pro-200K-Filtered}, and 
Wildchat \url{https://huggingface.co/datasets/allenai/WildChat-1M}\label{fn:instruct}. This model was subsequently replaced by Lucie-7B-Instruct-v1.1 due to better benchmark performance.} Because they have not learned basic skills like answering questions, it is hard to get a feel for model quality by interacting directly with a foundation model. The point of releasing the instruction models was thus to provide an illustration of how the base model could be fine-tuned to perform basic tasks with the idea that more complete fine-tuning and alignments would follow. 

Lucie-7B-Instruct-human-data is fine-tuned only on human-produced instructions collected either from open annotation campaigns or by applying templates to extant datasets, many of which were originally designed for NLP tasks. Lucie-7B-Instruct-v1.1 is fine-tuned on a mixture of templated data and synthetic instructions produced by ChatGPT. While the former performs less well on certain benchmarks than the latter, the interest of providing a model fine-tuned on human-created data is to give an example of what can be done to fine-tune LLMs to follow instructions without appealing to third party LLMs.

\subsection{Instruction datasets}
Instructions can be open-ended as in \ref{ex:open} or closed as in \ref{ex:closed}. They can also include additional input text on which the LLM is supposed to base its response as in summarization, information extraction or rephrasing. (Examples taken from the Alpaca-cleaned dataset.\footnote{\url{https://huggingface.co/datasets/yahma/alpaca-cleaned}}) 
\ex.\label{ex:open} Instruction: "Give three tips for staying healthy." 

\ex.\label{ex:closed} Instruction: "Find the capital of Spain." 

\ex.\label{ex:context} Instruction: "Rewrite the following sentence using active voice." ; Input: "The news report was read by the captain."

Selecting instruction datasets is difficult for a variety of reasons. Apart from potential problems of quality, we faced two major difficulties. The first was to select datasets that were consistent with our initiative to create fully open-source models that depend as little on less open, third party models as possible. Unfortunately, the majority of large-scale instruction datasets are generated with models such as ChatGPT or the Llama family. Because the use of such data was hard to avoid for large-scale instruction tuning, we made the decision to release two models: one model in which we attempted to avoid synthetic data (Lucie-7B-Instruct-human-data) and one that included synthetic data (Lucie-7B-Instruct-v1.1).

The second difficulty was to find instruction data in languages other than English, in particular, French, but also in the other natural languages on which Lucie-7B was trained: German, Italian and Spanish. Many non-English datasets are simply translations of English datasets, and often, they are automatic translations with little to no quality control. There have been serious efforts to create (originally) non-English datasets such as the Aya dataset and the Open Assistant dataset, both cited below, but the quantity of data in such corpora remains far smaller than what one can get with synthetic data.

\subsubsection{Non-synthetic datasets}\label{sec:datasets-nonsynth}
The line that we draw between synthetic and non-synthetic datasets is an imperfect one. Certain datasets, like Dolly, might be originally created by humans but then machine-translated into other languages. Other datasets that we have listed in the non-synthetic category might employ models for quality control, as in the original Croissant-Aligned data, or for automatic templating. The distinction that we wish to make here is one in which large, third party LLMs are used to generate responses to instructions (or even the instructions themselves) and those in which the basic content is created by humans.

\textbf{Aya Dataset}\\ \url{https://huggingface.co/datasets/CohereForAI/aya_dataset}\\
The Aya Dataset~\citep{singh2024aya} is a multilingual instruction fine-tuning dataset consisting of prompt/instruction pairs written by humans in the target language. The data was collected by Cohere For AI~\footnote{\url{https://cohere.com/research}} via the Aya Annotation Platform.\footnote{\url{https://aya.for.ai/}} We selected data for English, French, German, Italian, and Spanish. License: Apache 2.0.

\textbf{Croissant Aligned Instruct}\\ \url{https://huggingface.co/datasets/OpenLLM-France/Croissant-Aligned-Instruct}. \\ 
This dataset consists of French-English translation pairs from the CroissantLLM dataset (\url{https://huggingface.co/datasets/croissantllm/croissant_dataset_no_web_data/tree/main/aligned_36b}). The first sentence of each pair is prefixed with an instruction such as ``Please translate the following sentence into French''. We used 20K examples sampled randomly from 80K total. License: CC BY-SA 4.0.

\textbf{Dolly curated}\\ \url{https://huggingface.co/datasets/argilla/databricks-dolly-15k-curated-multilingual}. \\
This dataset collection~\citep{ouyang2022training} includes the original databricks-dolly-15k\footnote{\url{https://huggingface.co/datasets/databricks/databricks-dolly-15k}} dataset created by Databricks and consisting of human-generated English instruction/response pairs. It also includes machine-translations in a variety of languages. We took the English examples together with the translations for English, French, German and Spanish. License: CC BY-SA 3.0.

\textbf{ENS}\\ \url{https://huggingface.co/datasets/Gael540/dataSet_ens_sup_fr-v1}.\\
The ENS dataset is a French corpus of questions about the French higher education system (e.g., ``Quelle est la différence entre une université et une grande école ?'') with responses. License: Apache 2.0.

\textbf{FLAN v2 Converted}\\ \url{https://huggingface.co/datasets/ai2-adapt-dev/flan_v2_converted}. \\
This is a converted version of the FLAN collection~\citep{longpre2023flan}\footnote{\url{https://github.com/google-research/FLAN/tree/main/flan/v2}} extracted from \url{https://huggingface.co/datasets/Open-Orca/FLAN} by the Allen Institute for AI.\footnote{\url{https://allenai.org/}} The data is in English. License: CC BY 4.0. 

\textbf{Hard-coded prompts}: This is a set of customized instructions that provide information about OpenLLM and Lucie. The instructions are adapted from \url{https://huggingface.co/datasets/allenai/tulu-3-hard-coded-10x}.

\textbf{OpenAssistant Conversations 1}\\ \url{https://huggingface.co/datasets/OpenAssistant/oasst1}.\\
This dataset~\citep{köpf2023openassistantconversationsdemocratizing} is a crowd-sourced, multilingual conversation corpus. We used data in English, French, German, Italian and Spanish. License: Apache 2.0.

\textbf{Oracle}\\ 
\url{https://github.com/opinionscience/InstructionFr/tree/main/wikipedia}.\\
Oracle is a set of questions and answers based on French Wikipedia articles. It is a subset of InstructionFr, a repository of instruction datasets in French designed to finetune LLMs and released by Opsci. License: CC BY-SA.

\textbf{PIAF}\\ \url{https://huggingface.co/datasets/AgentPublic/piaf}.\\ 
PIAF is a reading comprehension dataset that includes answers based on pairs of questions and extracts from French Wikipedia articles \citep{keraron-EtAl:2020:LREC}. License: MIT.

\subsubsection{Synthetic datasets} \label{sec:datasets-synth}
\textbf{Alpaca-cleaned-fr}\\ \url{https://huggingface.co/datasets/cmh/alpaca_data_cleaned_fr_52k}. \\
Alpaca-cleaned-fr~\citep{vigogne} is built from the cleaned version of the Stanford Alpaca dataset~\citep{alpaca} and translated into French using gpt-3.5-turbo following the methods provided in \url{https://github.com/bofenghuang/vigogne/blob/main/docs/data.md}. License: CC BY-NC 4.0.

\textbf{Open Hermes 2.5}\\ \url{https://huggingface.co/datasets/teknium/OpenHermes-2.5}\\
A collection of diverse synthetic instruction datasets in English, totaling 1 million examples. Licenses: Mixed open.

\textbf{TULU3 Personas Math}\\ \url{https://huggingface.co/datasets/allenai/tulu-3-sft-personas-math}\\
This dataset was created using the personas method in order to improve model capacity to solve complex math word problems. License: ODC-BY. 

\textbf{TULU3 Personas Math Grade}\\ \url{https://huggingface.co/datasets/allenai/tulu-3-sft-personas-math-grade}\\
Like Personas Math, this dataset was created using personas with a focus on lower-level math word problems. License: ODC-BY.

\textbf{Wildchat}\\ \url{https://huggingface.co/datasets/allenai/WildChat-1M}.\\ 
WildChat~\citep{zhao2024wildchat} is a collection of 1 million conversations between human users and ChatGPT. We used the French subset. License: ODC-BY.\\

\subsection{The Instruct models}
When designing the data mix for Lucie-7B-Instruct-human-data, our approach was to use as much good quality human and templated data as possible in the five languages on which Lucie-7B was trained. When it came to mixes involving synthetic data, however, we tried many combinations before arriving at Lucie-7B-Instruct-v1.1,\footnote{Including Lucie-7B-Instruct, which was released simultaneously with Lucie-7B, but has since been replaced by Lucie-7B-Instruct-v1.1, as explained in footnote \ref{fn:instruct}.} which more than doubled the performance on GSM8K of the synthetic mix in second place. Interestingly, it did this while performing as well as or even better than the other mixes on French benchmarks despite the very small fraction of French instruction data used to train Lucie-7B-Instruct-v1.1.

\subsubsection{Data mixes}
Details on the individual datasets used in each data mix are given in Table \ref{tab:dataset_instruct_samples}. Table \ref{tab:dataset_instruct_categories} provides a breakdown of both mixes by type of data, synthetic or non-synthetic, and language. The statistics shown are those after two types of filtering. First, for the human data mixture, we filtered Aya Dataset, Dolly and Open Assistant to keep only languages on which Lucie-7B was trained. Second, even though the idea of the human data mix was to avoid synthetic data, as explained above, the distinction is imperfect and some synthetic data might have slipped through. We thus filtered both of our instruction mixes  for certain keywords expected for synthetic data; that is,  examples containing assistant responses were filtered out if the responses contained a keyword from the list \texttt{filter\_strings} in \ref{ex:filter}. This filter is designed to remove examples in which a generated response is presented as being from a model other than Lucie.

\ex.\label{ex:filter} \texttt{filter\_strings = ["OpenAI",
    "Open AI",
    "ChatGPT",
    "Chat GPT",
    "GPT-3",
    "GPT3",
    "GPT 3",
    "GPT-4",
    "GPT4",
    "GPT 4",
    "GPT-3.5",
    "GPT3.5",
    "GPT 3.5",
    "BingChat",
    "Bing Chat",
    "LAION",
    "Open Assistant",
    "OpenAssistant",
    "BARD",
    "PaLM",
    "Gemini",
    "Gemma",
    "Google AI",
    "Anthropic",
    "Claude",
    "LLaMA",
    "Meta AI",
    "Mixtral",
    "Mistral"]}

\begin{table}[h] 
\caption{
    The datasets with number of samples per language used for training Lucie-7B-Instruct-human-data (``human-data'') and Lucie-7B-Instruct-v1.1 (``v1.1''). 
}
\label{tab:dataset_instruct_samples}
\centering \begin{tabular}{llrr}
\toprule
    \textbf{Dataset} & \textbf{Language} & \textbf{\# human-data} & \textbf{\# v1.1}\\
    \midrule
    Alpaca fr & English & -- & 51,655 \\
    \midrule
    Aya Dataset & English & 3944 & -- \\
    & French & 1422 & -- \\
    & German & 241 & -- \\
    & Italian & 738 & -- \\
    & Spanish & 3854 & -- \\
    \midrule
    Dolly curated & English & 15,015 & -- \\
    & French & 15,015 & -- \\
    & German & 15,015 & -- \\
    & Spanish & 15,015 & -- \\
    \midrule 
    Croissant Aligned Instruct & French-English & 20,000 & 20,000 \\
    \midrule
    ENS & French & 394 & 394 \\
    \midrule 
    FLAN v2 Converted & English & 78,580 &  78,580\\
    \midrule
    Open Assistant 1 & English & 21,151 & -- \\
    & French & 1223 & -- \\
    & German & 1515 & -- \\
    & Italian & 370 & -- \\
    & Spanish & 14078 & -- \\
    \midrule
    Open Hermes & English & -- & 1,000,495 \\
    \midrule
    Oracle & French & 4613 & 4613 \\
    \midrule
    Personas Math & English & -- & 149,954 \\
    \midrule
    Personas Math Grade & English & -- & 49,964 \\
    \midrule 
    PIAF & French & 1849 & 1849 \\
    \midrule
    WildChat & French & -- & 26,436 \\    
    \midrule 
    Hard-coded prompts & English & 240 & 240 \\
     & French & 240 & 240 \\
     \bottomrule
\end{tabular}
\end{table}

\begin{minipage}{\textwidth}
    ~
\end{minipage}

\begin{table}[H] 
\caption{
    A breakdown of the data mixes for Lucie-7B-Instruct human data (``human-data'') and Lucie-7B-Instruct-v1.1 (``v1.1'') by category. Synthetic data comes from the datasets listed in Section \ref{sec:datasets-synth}; non-synethetic from those in \ref{sec:datasets-nonsynth}. Language is determined by the principal language listed for the original dataset.
}
\label{tab:dataset_instruct_categories}
\centering \begin{tabular}{lrr}
\toprule
    \textbf{Category} & \textbf{\# human-data} & \textbf{\# v1.1}\\
    \midrule
    Synthetic &  -- & 1,278,504 \\
    Non-synthetic  & 214,512& 105,916 \\
    \midrule
    English & 118,930 & 1,279,233 \\
    French & 24,756 & 85,187 \\
    Translated data & 20,000 & 20,000 \\
    Other (de, es, it)  & 50,826 & -- \\
    \bottomrule
\end{tabular}
\end{table}

\subsubsection{Training procedure}
    
The model architecture and hyperparameters are the same as for Lucie-7B during the annealing phase with two exceptions. First, we use a context length of 4096 tokens and a batch size of 1024 (as in the main pretraining phase), though we check that the model can still handle context sizes of up to 32,000 tokens (see Section~\ref{sec:evaluation_niah}). Second, we use a cosine learning rate scheduler from 3e-5 to 3e-6.

We use a chat template similar to Llama 3.1:
\begin{tcolorbox}[colback=gray!10]
\begin{verbatim}
<s><|start_header_id|>user<|end_header_id|>

Donne trois conseils pour rester en bonne santé.

<|eot_id|><|start_header_id|>assistant<|end_header_id|>

1. Mangez une alimentation équilibrée et assurez-vous d'inclure beaucoup de fruits 
et légumes. 
2. Faites de l'exercice régulièrement pour maintenir votre corps actif et fort. 
3. Dormez suffisamment et maintenez un horaire de sommeil régulier.

<|eot_id|>
\end{verbatim}
\end{tcolorbox}

As such, we introduce three new special tokens:
\begin{itemize}
    \item \textbf{\texttt{<|start\_header\_id|>}} and \textbf{\texttt{<|end\_header\_id|>}}: These tokens circumscribe the role of the speaker, which can either be: system, user or assistant.
    \item \textbf{\texttt{<|eot\_id|>}}: This token marks the end of a turn in the conversation.
\end{itemize}

Unlike the pretraining phase, we do not use data packing. Instead, each sample is truncated or padded to a fixed length of 4096 tokens. Additionally, gradient descent is applied only to the assistant's responses.

\subsection{Evaluation}\label{sec:evaluation_niah}

\subsubsection{Benchmark evaluations}

We evaluate the instruction models on the same categories of benchmarks detailed in Section \ref{sec:bench-pretraining}. We also add results for French generative benchmarks from \cite{croissant}: FQuad, a French Question Answering dataset containing manually annotated sets of Wikipedia passages (FQuADGenQ and MultiFQuAD); OrangeSum, \footnote{\url{https://huggingface.co/datasets/EdinburghNLP/orange_sum}} a summarization dataset of French news articles; and French Trivia, \footnote{\url{https://huggingface.co/datasets/manu/french-trivia}} a dataset containing questions on French culture, written in English.

\begin{figure}[h]
    \centering
    \includegraphics[width=\textwidth]{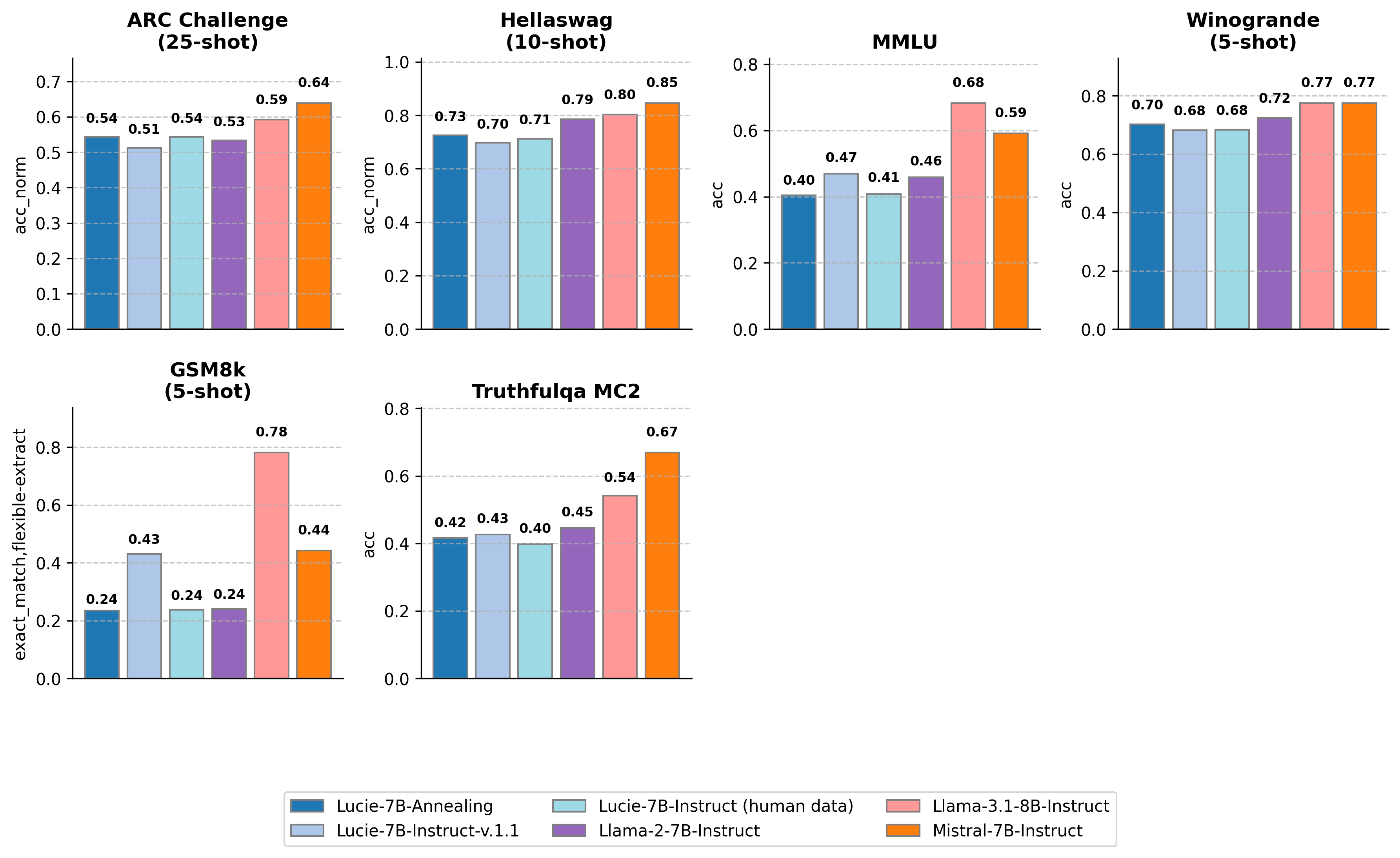}
    \caption{
        Comparative results of the Lucie instruct models on English NLP benchmarks.
    }
    \label{fig:instruct_c}
\end{figure}

\begin{figure}[h]
    \centering
    \includegraphics[width=\textwidth]{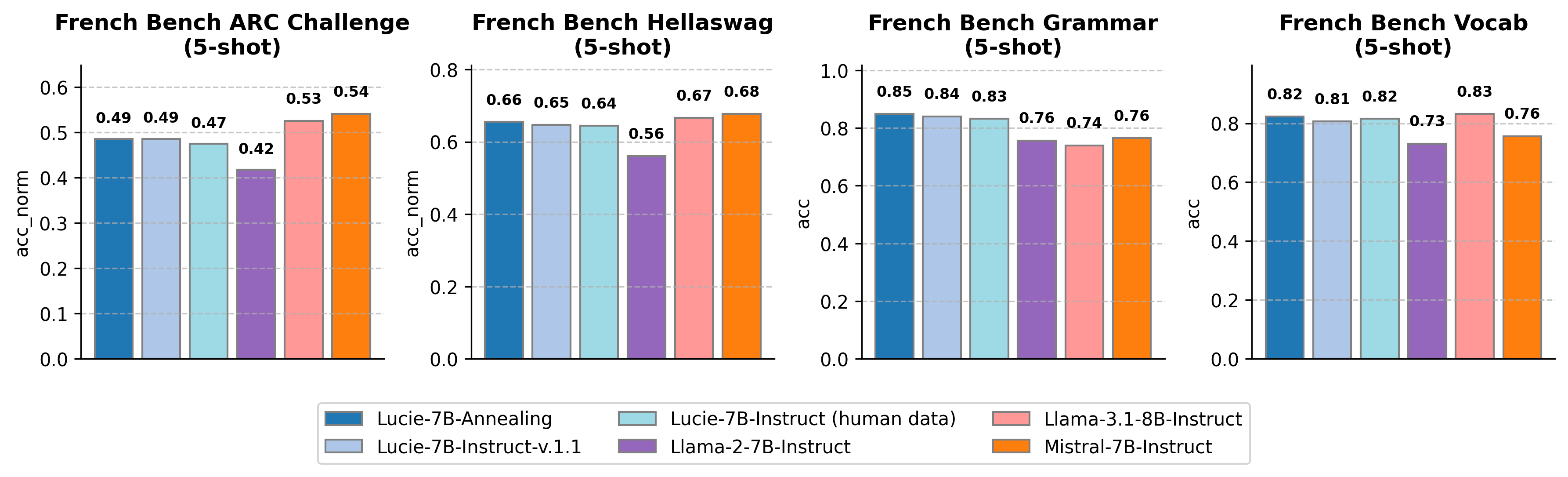}
    \caption{
    Comparative results of the Lucie instruct models on French multi-choices benchmarks.
    }
    \label{fig:instruct_b}
\end{figure}

\begin{figure}[h]
    \centering
    \includegraphics[width=\textwidth]{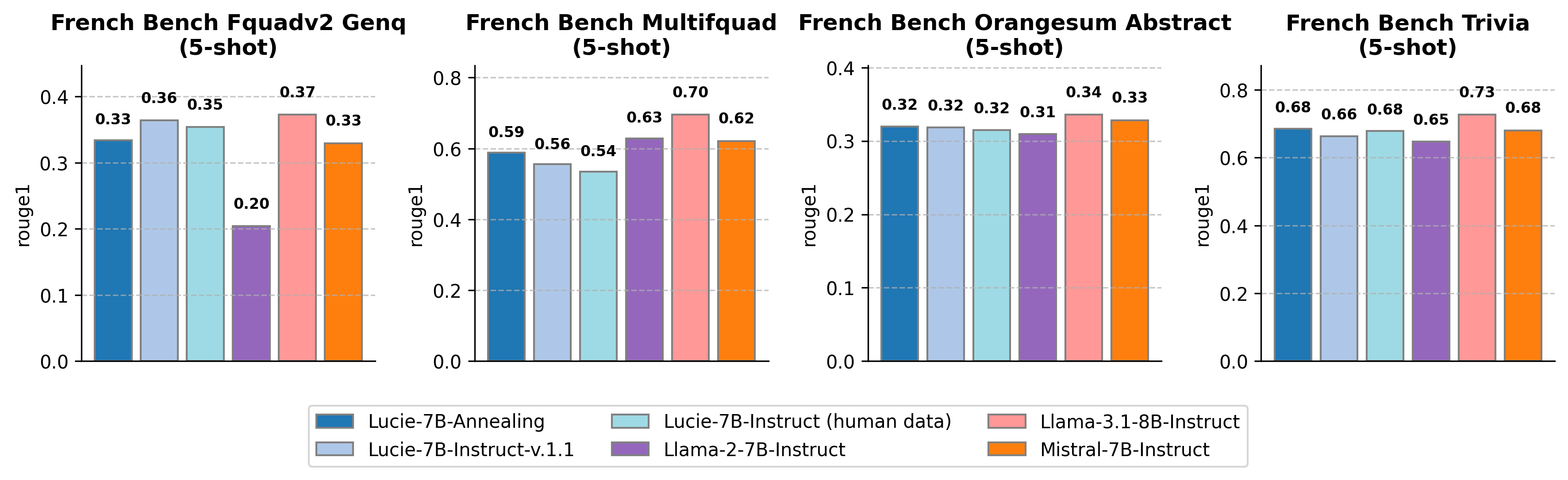}
    \caption{
        Comparative results of the Lucie instruct models  on French generative benchmarks.
    }
    \label{fig:instruct_a}
\end{figure}

\begin{figure}[h]
    \centering
    \includegraphics[width=\textwidth]{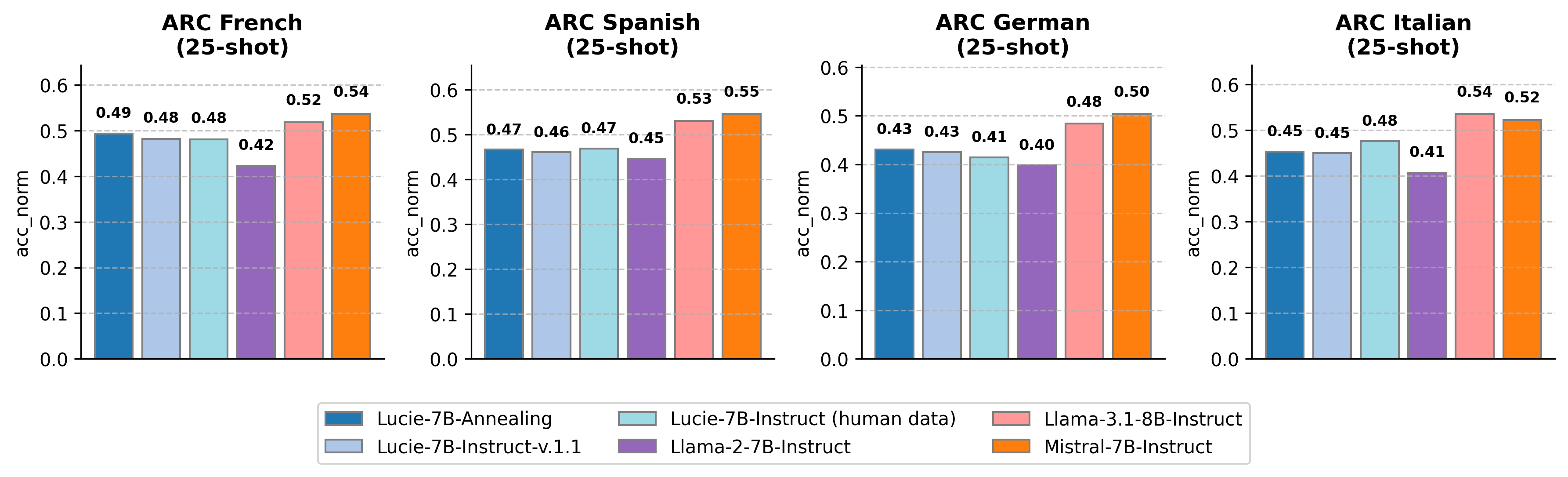}
    \caption{
        Comparative results of the Lucie instruct models  on multilingual benchmarks.
    }
    \label{fig:instruct_d}
\end{figure}

Results for the Lucie instruct models are given in Figures \ref{fig:instruct_c}, \ref{fig:instruct_b},  \ref{fig:instruct_a} and \ref{fig:instruct_d}. We add Lucie-7B (after annealing), Llama-2-7B-Instruct, Llama-3-8B-Instruct and Mistral-7B-Instruct for comparison.

The biggest improvement, at least for Lucie-7B-Instruct-v1.1, comes on GSM8K, where we see a nearly 20 point increase relative to Lucie-7B after annealing, Llama-2-7B-Instruct and also Lucie-7B-Instruct-human-data. While results still fall below that of Llama-3-8B-Instruct, Lucie's performance is comparable to Mistral-7B-Instruct.

Interestingly, Lucie-7B-Instruct-v1.1 manages to keep the relatively strong performance of Lucie-7B on French benchmarks despite being trained on very little French instruction data (around 6\%). While slightly outperforming Lucie-7B-Instruct-human-data on some generative benchmarks, the two Lucie instruct models are roughly equivalent in multilingual performance.

Results on English benchmarks still fall clearly below the state of the art. Here again, future work will involve looking at how dataset composition impacts this performance, but given the gap already present for the pretrained models, the discrepancy between the Lucie instruct models and other comparison models likely stems from the pretraining stage.

\subsection{Needle in a haystack}
\begin{figure}[htp]
    \centering
    \begin{minipage}{0.45\textwidth}
        \centering
        \includegraphics[width=\textwidth]{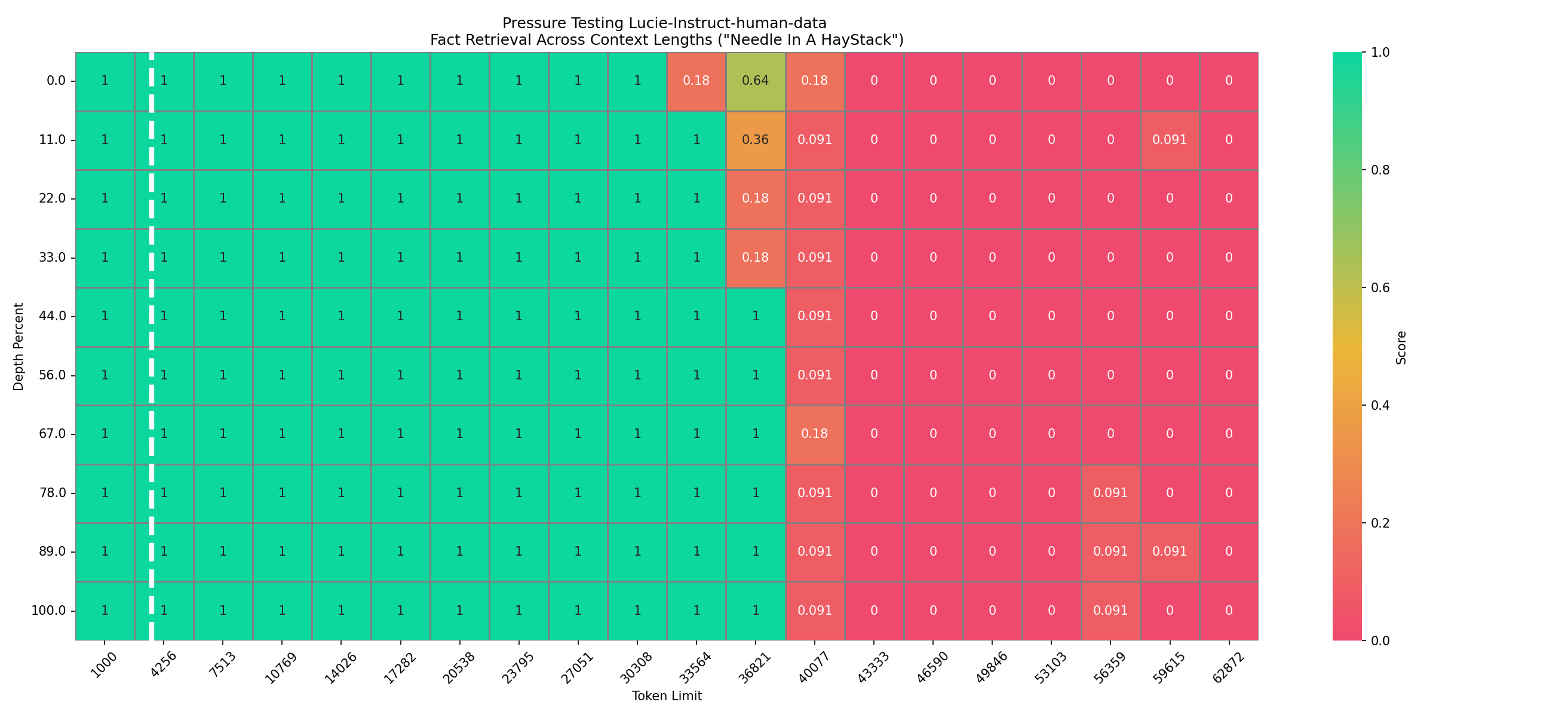}
        \caption{
            Needle in a haystack results on Lucie-7B-Instruct human-data.
        }
        \label{fig:inst_needle_a}
    \end{minipage}
    \hfill
    \begin{minipage}{0.45\textwidth}
        \centering
        \includegraphics[width=\textwidth]{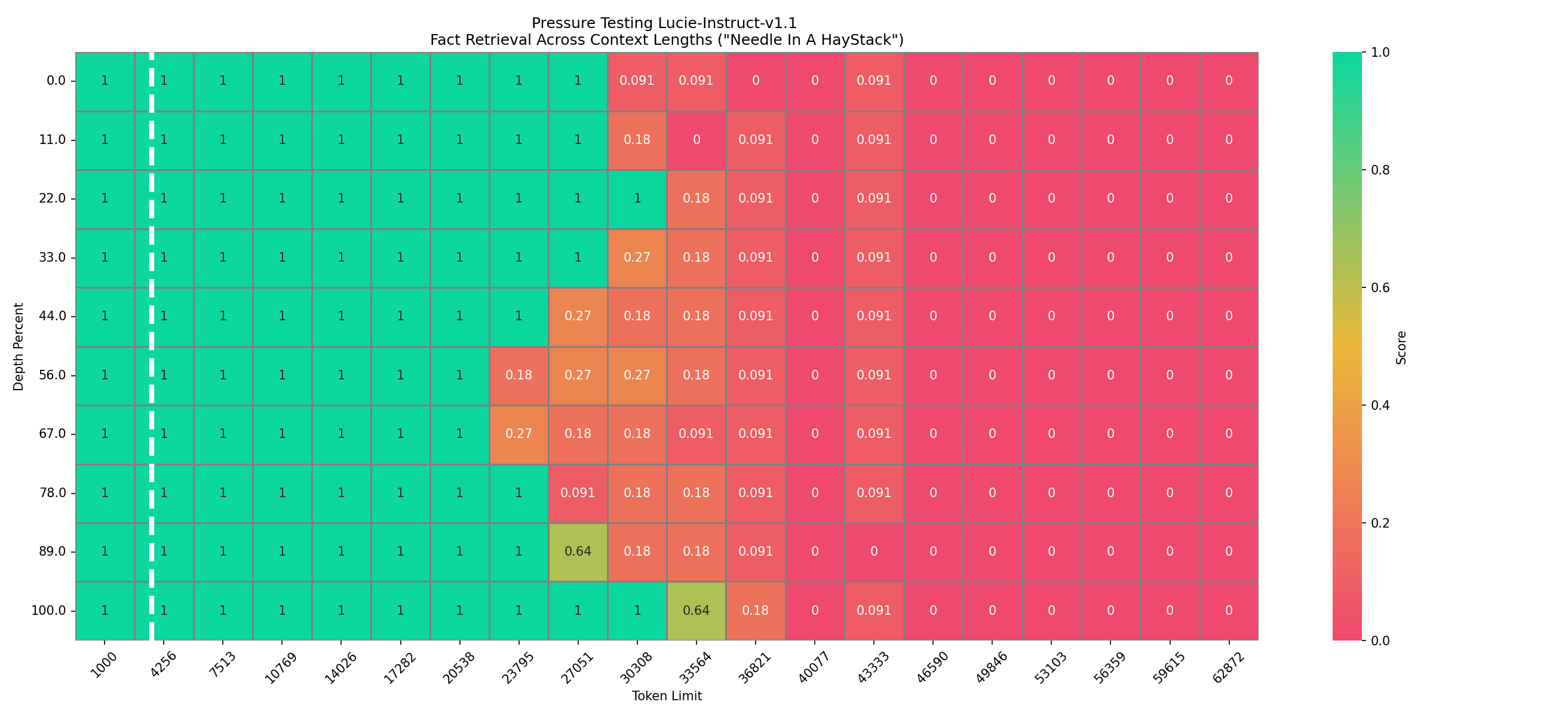}
        \caption{
            Needle in a haystack results on Lucie-7B-Instruct-v1.1.
        }
        \label{fig:inst_needle_b}
    \end{minipage}
\end{figure}

Evaluation of Lucie-7B-Instruct-human-data on the ``Needle in the Haystack'' benchmark (see Section \ref{sec:needle-pretraining}) shows no loss on context window sizes up to 32,000 tokens compared to the pretrained model. The window falls considerably, to 22,000 tokens, for Lucie-7B-Instruct-v1.1, however. We suspect that this reduction can be attributed to the fact that Lucie-7B-Instruct-v1.1 is trained on significantly more data than Lucie-7B-Instruct-human-data and that instruction fine-tuning was carried out, for both models, on sequences of 4096 tokens.\footnote{We note that the original Lucie-7B-Instruct model maintained a context window length of 32,000 tokens even though it, too, was trained on sequences of 4096 tokens. It was trained on about a quarter of the data of Lucie-7B-Instruct-v1.1.} The results are shown in Figures~\ref{fig:inst_needle_a} and~\ref{fig:inst_needle_b}.

The reduction in context window size exhibited by Lucie-7B-Instruct-v1.1 underscores the need for fine-tuning and alignment of the Lucie-7B foundation model with datasets including longer training sequences. As the point of the first instruct models was to provide preliminary illustrations of Lucie-7B's potential, we leave such training for future work.


\section{Conclusion}

We publicly release the Lucie Training Dataset and the Lucie-7B family of LLMs in an effort to demonstrate the feasibility, and better understand the impact, of making best efforts to comply with regulation and move away from US-centric biases when selecting data and training a new sovereign LLM from scratch. The Lucie Training Dataset is particularly well adapted to French use cases, containing nearly 600 billion  words in French from a variety of documents with unrestrictive licenses, but also contains data in English, German, Italian, Spanish and code. Lucie-7B was trained on equal amounts of English and French data to offset biases introduced by training on English-heavy datasets. After going through a context length extension phrase during pretraining, the model has a context length of 32,000 tokens. To improve its performance on selected benchmarks, we carried out at the very end of training an annealing phase in which we added high quality data focused on math and natural language processing tasks from FLAN v2. We note however that, although this annealing phase clearly improves mathematical reasoning, it may have been better to focus more on math and code earlier on in the training process. We could likely also benefit from the use of more data, as well as the use of reasoning elicitation approaches such as long chain-of-thought fine-tuning.

While our principal focus has been on the training dataset and foundation model, we also release two instruction tuned models as illustrations of how the model can behave with simple fine-tuning. We publish two versions: one version containing only data templated or otherwise created by humans and one version containing synthetic data produced by ChatGPT. In the coming months our focus will turn to aligning Lucie-7B with preference data to further improve the model's behavior and adapt it to an educational context.


\section{Acknowledgements}

This work was performed using HPC resources from GENCI–IDRIS (Grant 2024-GC011015444). We gratefully acknowledge funding support from GENCI and IDRIS and from Pierre-François Lavallée (IDRIS) and Stephane Requena (GENCI) in particular.

We also thank the support teams from IDRIS, in particular Myriam Peyrounette and Hatim Bourfoune, and from Hugging Face, in particular Thomas Wolf, Guilherme Penedo, Elie Bakouch, Haojun Zhao, and Lucain Pouget for their technical guidance throught the project.

\textbf{Dataset}

The Lucie Training Dataset was created by members of LINAGORA (Olivier Gouvert, Julie Hunter, Jérôme Louradour, Jean-Pierre Lorré) and the OpenLLM-France community.

We thank in particular Rachel Bawden (INRIA), Clément Bénesse (Opsci), Christophe Cérisara (LORIA), Olivier Ferret (CEA List), Joöl Gombin (Opsci), Ismaïl Harrando (LINAGORA), Jordan Ricker (Opsci), Guokan Shang (MBZUAI), and Yaya Sy (LORIA) for their helpful input.

\textbf{Tokenizer}

The Lucie tokenizer was trained by Jérôme Louradour (LINAGORA) with input from (in alphabetical order): Rachel Bawden (INRIA), Christophe Cerisara (LORIA), Evan Dufraisse (CEA List), Manuel Faysse (ILLUIN Technology), Ismaïl Harrando (LINAGORA/SciencesPo), 
 Julie Hunter (LINAGORA), Jean-Pierre Lorré (LINAGORA), Michel-Marie Maudet (LINAGORA), Guokan Shang (MBZUAI), and Yaya Sy (LORIA).

\textbf{Lucie-7B}

Lucie-7B was created by members of LINAGORA and the OpenLLM-France community, including in alphabetical order: Thibaut Boissin (IRT), Christophe Cerisara (LORIA), Evan Dufraisse (CEA List), Olivier Gouvert (LINAGORA), Lucas Hervier (IRT), Julie Hunter (LINAGORA), Jean-Pierre Lorré (LINAGORA), Jérôme Louradour (LINAGORA), Michel-Marie Maudet (LINAGORA), Agustin Martin Picard (IRT)  and Yaya Sy (LORIA).

We thank Rachel Bawden (INRIA), Clément Bénesse (Opsci), Manuel Faysse (ILLUIN Technology), Olivier Ferret (CEA List), Joël Gombin (Opsci), Ismaïl Harrando (LINAGORA/SciencesPo), Jordan Ricker (Opsci), Guokan Shang (MBZUAI), and Julien Tourille (CEA List) for their helpful input.

\textbf{Lucie instruct models}

Lucie-7B was created by members of LINAGORA and the OpenLLM-France community, including in alphabetical order: Olivier Gouvert (LINAGORA), Ismaïl Harrando (LINAGORA/SciencesPo), Julie Hunter (LINAGORA), Jean-Pierre Lorré (LINAGORA), Jérôme Louradour (LINAGORA), Michel-Marie Maudet (LINAGORA), and Laura Rivière (LINAGORA).

We thank Clément Bénesse (Opsci), Christophe Cerisara (LORIA), Evan Dufraisse (CEA List),  Olivier Ferret (CEA List), Joël Gombin (Opsci), Émile Hazard (Opsci), Jordan Ricker (Opsci) and Guokan Shang (MBZUAI), and for their helpful input.

Finally, we thank the entire OpenLLM-France community, whose members have helped in diverse ways.

\bibliography{biblio}

\end{document}